\documentclass[letterpaper]{article} 
\usepackage[]{aaai23} 
\usepackage{times}  
\usepackage{helvet}  
\usepackage{courier}  
\usepackage[hyphens]{url}  
\usepackage{graphicx} 
\usepackage{color}
\usepackage{float}
\usepackage{natbib}  
\usepackage{caption} 
\usepackage{algorithm}
\usepackage{algpseudocode}
\usepackage{newfloat}
\usepackage{listings}
\usepackage{bbm}
\usepackage{mathtools}
\usepackage{amsmath}
\usepackage{amssymb}
\usepackage{amsthm}
\usepackage{wrapfig}

\usepackage{microtype}
\newtheorem{theorem}{Theorem}[section]

\newcommand{\colorb}[1]{\textcolor{blue}{#1}}
\usepackage{microtype}
\usepackage{graphicx}
\usepackage{subfigure}
\algnewcommand\algorithmicforeach{\textbf{for each}}
\algdef{S}[FOR]{ForEach}[1]{\algorithmicforeach\ #1\ \algorithmicdo}

\DeclareCaptionStyle{ruled}{labelfont=normalfont,labelsep=colon,strut=off} 
\floatstyle{ruled}
\newfloat{listing}{tb}{lst}{}
\floatname{listing}{Listing}
\title{Decision-making with Speculative Opponent Models}
\author{
    Jing Sun\textsuperscript{\rm 1},
    Shuo Chen\textsuperscript{\rm 2}\thanks{Corresponding author},
    Cong Zhang\textsuperscript{\rm 1},
    Yining Ma\textsuperscript{\rm 1},
    Jie Zhang\textsuperscript{\rm 1}
}
\affiliations{
    \textsuperscript{\rm 1}School of Computer Science and Engineering, Nanyang Technological University\\

    Singapore 639798\\
    \textsuperscript{\rm 2}Beijing Institute for General Artificial Intelligence,
\\

    Beijing, China 100091\\
    
    \{sun.jing, yining.ma, zhangj\}@ntu.edu.sg, cong030@e.ntu.edu.sg, 
    chenshuo@bigai.ai
}

\usepackage{bibentry}

\begin{document}

\maketitle

\begin{abstract}
Opponent {modelling} has proven effective in enhancing the decision-making of the controlled agent by constructing models of opponent agents. However, existing methods often rely on access to the observations and actions of opponents, a requirement that is infeasible when such information is either unobservable or challenging to obtain. {To address this issue, we introduce Distributional Opponent-aided Multi-agent Actor-Critic (DOMAC), the first speculative opponent {modelling} algorithm that relies solely on local information (i.e., the controlled agent's observations, actions, and rewards). Specifically, the actor maintains a speculated belief about the opponents using the tailored \textit{speculative opponent models} that predict the opponents' actions using only local {information}. Moreover, DOMAC features distributional critic models that estimate the return distribution of the actor's policy, yielding a more fine-grained assessment of the actor's quality. This thus more effectively guides the training of the speculative opponent models that the actor depends upon. Furthermore, {we formally derive a policy gradient theorem with the proposed opponent models.}
Extensive experiments under eight different challenging multi-agent benchmark tasks within the MPE, Pommerman and StarCraft Multiagent Challenge (SMAC) demonstrate that our DOMAC successfully models opponents' {behaviours} and delivers superior performance against state-of-the-art methods with a faster convergence speed.}
\end{abstract}

\section{Introduction}\label{sec:intro}
{R}{ecently}, there has been a growing effort in applying multi-agent reinforcement learning (MARL) to address the complex learning tasks in multi-agent systems, including cooperation, competition, {and a mix of both~\cite{hernandez2019survey,zhang2021multi,yu2021review, papoudakis2020benchmarking}. The framework of centralized training with decentralized execution (CTDE) has drawn enormous attention, {where the policy of each agent is trained} with access to global information in a centralized way and {executed} given only the local observations in a decentralized {manner}. Empowered by CTDE, {plenty of multi-agent RL (MARL) methods are proposed~\cite{lowe2017multi,foerster2018counterfactual, rashid2018qmix, sunehag2017value, yu2022surprising}, including policy gradient-based and value-based ones.}
The policed-based methods, such as MADDPG\cite{lowe2017multi}, COMA \cite{foerster2018counterfactual} and MAPPO \cite{yu2022surprising} use the information (observations and actions) of the controlled agents to train the team policy. The value-based methods {on the other hand}, {with} VDN \cite{sunehag2017value} and QMIX \cite{rashid2018qmix} {as the exemplars}, 
construct a joint value function for the controlled agents.
These works normally regard the opponents (if exist) as a part of the environment. 

However, {some} people argue that the controlled agent should be endowed with abilities to reason about the opponents' unknown goals and {behaviours} {to benefit the decision-making}, thus coming up with opponent {modelling}~\cite{albrecht2018autonomous}. By successfully {modelling} the opponents, the agent can {reliably} estimate opponents' {behaviours} as well as their goals, and adjust its policy to achieve optimal decision-making.} {Several recent works have proposed learning opponent models using deep learning architectures \cite{he2016opponent,foerster2018learning,raileanu2018modeling,tian2019regularized,yang2018towards}}.

{Nonetheless}, the {existing} works commonly assume access to the opponents' observations and actions during {\textit{both}} training and execution. Recent
works {relax the requirement of} access to opponents' observations and actions during execution \cite{papoudakis2021agent,papoudakis2020local}. {However, they still assume} the opponents' {information} as the ground truth for training {the} opponent models. 
{We observe that} access to {the} opponents' real observations and actions {during training} may not be available or cheap in many cases. For instance, even in a simulator {where} the user has complete control, the opponent's policy can come from third-party resources, and thus, its observation format is unknown. When it comes to realistic training environments, {although} we have the full knowledge of the opponents' configuration, collecting the opponents' running data induces high costs as the number of agents and task complexity increase. In these cases, the controlled agent can only rely on its locally available information (i.e., its own observations, actions, and rewards) to model its opponents if it wants to benefit from opponent {modelling}, which seems counter-intuitive. Thus, a {natural but challenging} question is:
{Can} we portray the opponents' {behaviours} with only the controlled agent's \textit{local} information? 

{{To answer this question}, this work aims to estimate the opponents' {behaviours} with {\textit{only}} the controlled agent's local information.} 
For this purpose, we design a novel opponent modelling workflow. That is, the controlled agent first speculates its opponents' behaviours using opponent models conditioned on its local information; Then, the controlled agent makes decisions based on its speculation. Due to this design, the agent's decisions are deeply coupled with the opponent models. We expect that a better opponent model can improve the decision quality because it provides the controlled agent with more accurate information about the opponents. Therefore, it is possible to use the return resulting from the controlled agent's decisions as one kind of feedback, which takes the place of the true opponent information, to train the opponent models.
This is the key insight of our work.

Following the above workflow, we propose the Distributional Opponent-aided Multi-agent Actor-Critic (DOMAC), an innovative algorithm tailored for speculative opponent modelling utilizing solely local information. The approach consists of two primary modules: the Opponent-aided Actor (OMA) and the Centralized Distributional Critic (CDC). 
Within the OMA, the controlled agent, i.e., the actor, maintains one opponent model for each opponent, which predicts the opponent's actions based on the controlled agent's local information.
Although these inputs are {not directly from the} opponent strategies, they still {correlate} with the opponents' {behaviours}. For example, the opponents {could} appear in the observations of the controlled agents, {thus} the local observations can occasionally convey the state-changing information of the opponents.
Since our opponent modelling does not use the opponents' information as the ground truth during training or execution, we call our opponent models the \emph{speculative opponent models}.
These speculative opponent models are integral to enhancing the actor's decision-making capabilities by providing refined predictions for improved performance.
Meanwhile, to gather as much feedback as possible for training the OMA,
the CDC capitalizes on the concept of distributional critics ~\cite{bellemare2017distributional,dabney2018implicit,dabney2018distributional,bellemare2023distributional,lyu2020likelihood,hu2020qr,sun2021dfac, huang2022distributional} to appraise the distribution of the actor's returns, rather than the conventional scalar estimate. This distributional perspective allows for the explicit consideration of correlations between values, thereby capturing more information concerning the return. 
It avails a more nuanced and informative evaluation of the actor's policy that underpins the refinement of the speculative opponent models essential for the actor's optimization.
We highlight the following contributions: 

\begin{itemize}
    \item

To the best of our knowledge, we are the first work that studies the opponent {modelling} with purely local information from the controlled agent. 
\item  {In the proposed DOMAC algorithm, we have tailored the designs of OMA and CDC. Besides, we are pioneering the use of distributional critics to guide the training of the opponent models, which can inspire future research.
\item Furthermore, we present a formal derivation, grounded in the policy gradient theorem, that guides the joint training of both our agent policy and opponent models, bolstering the theoretical foundation of our DOMAC designs.
\item  Lastly, we empirically show that DOMAC significantly outperforms existing state-of-the-art methods on several challenging multi-agent tasks, including MPE\cite{lowe2017multi}, Pommerman\cite{DBLP:journals/corr/abs-1809-07124}, and StarCraft Multiagent Challenge (SMAC) \cite{samvelyan2019starcraft}. The extensive experiments confirm that our method successfully learns reliable opponent models without using the opponents' true information and achieves better task performance with a faster convergence speed than baselines.}
\end{itemize}

{The remainder of this paper is organized as follows. Section \ref{s2} briefly reviews the works in multi-agent RL, opponent {modelling}, and distributional RL. Section \ref{s3} introduces the preliminaries of the proposed method. Section \ref{s4} elaborates on the detailed designs of the DOMAC. Section \ref{s5} provides the evaluation experiments and ablations studies. Finally, Section \ref{s6} concludes the paper and presents future works.}

\section{Related work}\label{s2}
\noindent\textbf{Multi-agent reinforcement learning. }
In recent years, the developments in multi-agent reinforcement learning (MARL) have led to great progress in creating artificial agents that can efficiently cooperate to solve tasks~\cite{hernandez2019survey, zhang2021multi, gronauer2022multi}. MARL algorithms generally fall into two frameworks: centralized and decentralized learning. Centralized methods \cite{claus1998dynamics} learn a single policy to produce the joint actions of all agents directly, while decentralized learning \cite{littman1994markov} entails each agent independently optimizing its reward. The framework of centralized training and decentralized execution (CTDE) bridges the gap between the two aforementioned frameworks, which permits the sharing of information during training, while policies are only conditioned on the agents' local observations enabling decentralized execution\cite{lowe2017multi}. One category of CTDE algorithm is policy gradient methods, wherein each agent consists of a decentralized actor and a centralized critic, which is optimized by the shared information of the controlled agents\cite{lowe2017multi,foerster2018counterfactual}. The value decomposition method
is another category, which represents the joint Q-function as a function of agents' local Q-functions \cite{sunehag2017value, rashid2018qmix, son2019qtran,wang2020qplex}. However, these studies often consider opponents (if exist) as part of the environment, failing to explicitly model the influence of the opponents, which yields sub-optimal learning outcomes.

\noindent\textbf{Opponent {modelling}.}
 Opponent {modelling} is a research topic that emerged alongside the game theory~\cite{brown1951iterative}. With the powerful representation capabilities of recent deep learning architectures, opponent {modelling} has ushered in significant progress~\cite{albrecht2018autonomous}. 

Type-based reasoning methods~\cite{he2016opponent,hong2018deep,raileanu2018modeling,albrecht2017reasoning} assume that the opponent has one of several known types and update the belief using their observations obtained during training. 
Recursive reasoning methods ~\cite{yang2019towards, wen2019probabilistic,wen2019modelling} model the beliefs about the mental states of other agents via deep neural networks. 
The opponent {modelling} with online reasoning refers to the online inference of the opponent's policy through Bayesian inference and formulating a corresponding response \cite{zheng2018deep, tian2019regularized, digiovanni2021thompson, fu2022greedy}. 
The process of {modelling} opponents with dynamic strategies involves estimating the opponents' {behaviours} and corresponding impacts by using the opponent's learning \cite{foerster2017learning,grover2018learning,kim2021policy,liu2019multi}. The opponent modelling with meta-learning is designed to train against a set of known opponent policies to quickly adapt to unknown opponent policies during the testing \cite{zintgraf2021deep,wu2022l2e}. Theory of mind-based opponent {modelling} reasons about opponent's mental status and intentions, to predict and adapt to opponent {behaviour} by {modelling} their beliefs, goals, and actions \cite{yang2018towards,rabinowitz2018machine}. 
However, the aforementioned works commonly assume access to the opponents' observations and actions during both training and execution. Recent {research ~\cite {papoudakis2020local,papoudakis2021agent} argues} that access to opponents' observations and actions during execution is often infeasible (e.g., in large-scale applications). They manage to learn an opponent model that only uses the local information of the agent, such as its observations, actions, and rewards during execution. {Nevertheless, the efficacy of their methods is constrained by the necessity of accessing opponents' genuine data during the training phase, a requirement that significantly restricts their applicability in situations where such information is unobservable.} Thus, how to model an opponent's policy when its information is unavailable during training is still an open {research} problem. Our work presents the first attempt to solve this {challenge}. %

\noindent \textbf{Distributional reinforcement learning.}
Distributional reinforcement learning (RL) aims to model the return distributions rather than the expected return and uses these distributions to evaluate and optimize a policy ~\cite{bellemare2017distributional,dabney2018implicit,dabney2018distributional,bellemare2023distributional}. Many studies have shown that distribution RL can achieve better performance than the state-of-art methods from the classical RL ~\cite{barth2018distributed,tessler2019distributional,singh2020sample,yue2020implicit, bai2023pacer,zhang2023distributional,wiltzer2022distributional}. The {Categorical DQN (C51) \cite{bellemare2017distributional} employed a discrete set of $N$ fixed values to approximate the return distribution, achieving superior performance than DQN. Then, Dabney et al. introduced QR-DQN \cite{dabney2018distributional}, by appealing to the principle of quantile regression (QR) \cite{koenker2001quantile}, which assigns $N$ fixed and uniform probabilities to adjustable values. Furthermore, an extension to QR-DQN was realized through implicit quantile networks (IQNs) \cite{dabney2018implicit}, which learn the entire quantile function via neural networks. {Interestingly}, the latest research in Nature also shows that similar distributional mechanisms exist in human brains \cite{dabney2020distributional}, {motivating a variety of hybrid methodologies to blend these distributional methods with existing RL techniques to address issues within single-agent~\cite{barth2018distributed, ma2020dsac} and multi-agent scenarios~\cite{lyu2020likelihood,hu2020qr,sun2021dfac, huang2022distributional}.}} 

Our work {is the first to investigate distributional RL for opponent modelling}. Together with previous works, we show the great potential of distributional RL and can attract more research efforts to this area.

\section{Preliminary}\label{s3}
\noindent\textbf{Partially observable Markov games.} 
A \textit{partially observable Markov game} (POMG)~\cite{lowe2017multi} of $n$ agents is formulated as a tuple $\mathcal{M} = \langle \mathcal{S}, \mathbb{O}, \mathcal{O},  \mathcal{A}, \mathcal{T}, \mathcal{R}, \gamma \rangle$. $\mathcal{S}$ is a set of states describing the possible configuration of all agents and the external environment. Also, each agent $i$ has its own observation space $\mathbb{O}_i \in \mathbb{O}$. Due to the \textit{partial observability}, in every state $s \in \mathcal{S}$, each agent $i$ gets a correlated observation $o_i$ based on its observation function $\mathcal{O}_i: \mathcal{S}\rightarrow \mathbb{O}_i$ where $\mathcal{O}_i \in  \mathcal{O}$. The agent $i$ selects an action $a_i \in A_i$ from its own action space $A_i \in \mathcal{A}$ at each time step, giving rise to a joint action $[a_1, \cdots, a_n] \in A_1 \times A_2 \times \cdots \times A_n$. The joint action then produces the next state by following the state transition function $\mathcal{T}: \mathcal{S} \times A_i \times \cdots \times A_n \rightarrow \mathcal{S}$. {$\mathcal{R}=\{r_i\}$ is the set of reward functions}. After each transition, agent $i$ receives a new observation and obtains a scalar reward as a function of the state and its action $r_i: \mathcal{S} \times A_i \rightarrow \mathbb{R}$. The initial state $s \in \mathcal{S}$ is determined by some prior distribution $p: \mathcal{S} \rightarrow [0, 1]$. Each agent $i$ aims to maximize its own total expected return {$R_i = \mathbb{E}_{r^t\sim r_i(s_t, a_i^t), (s_t, a_i^t)\sim\tau} \sum_{t=0}^T \gamma^t r^t$, where $\gamma$ is the discount factor, $r^t$ is its sampled reward at time step $t$, $\tau$ is the trajectory distribution induced by the joint policy of all agents, and $T$ is the time horizon.} Without loss of generality, we assume that the $n$ agent can be divided into $|M| \leq n$ teams, and each team has $q$ agents with $1\leq q \leq n$. We consider the teams out of our control as opponent agents with {static policies}. Note that a single agent can also form a team. We assume agents from the same team fully cooperate and thus share the same reward function.

\noindent \textbf{Actor-Critic method.} The actor-critic algorithm ~\cite{konda2000actor} is a subclass of policy gradient methods ~\cite{silver2014deterministic}. It has been widely used for tackling complex tasks. The algorithm consists of two components. The $critic$ $Q_\phi^\pi$ estimates the true action-value function $Q^\pi(s, a)$ that represents the expected return of taking action $a$ in state $s$ and then following policy $\pi$. It adjusts the parameters $\phi$ with an appropriate policy evaluation algorithm such as temporal-difference learning \cite{sutton2018reinforcement}. With the critic, the $actor$ adjusts the parameter $\theta$ of the agent's policy $\pi_\theta$ by applying the policy gradient theorem \cite{sutton2000policy}:
\begin{equation}
    \nabla_\theta J(\theta) = \mathbb{E}_{\pi_\theta} [ \nabla_\theta \log \pi_\theta(a|s) Q_\phi^\pi(s, a) ].
    \label{eq1a}
\end{equation}
In multi-agent scenarios, the actor-critic algorithm is applied to learn the optimal policy for each agent. To stabilize the training, it is common to adopt the centralized training and decentralized execution (CTDE) setting \cite{hernandez2019survey}, where the agent $i$'s action-value function $Q_i^{\pi_i}(\mathbf{o}, \mathbf{a})$ takes as input the joint observation and joint action of the agents in the same controlled team. {The gradient {of the policy for agent $i$ is given as:}}
\begin{equation}
    \nabla_{\theta_i} J(\theta_i) = \mathbb{E}_{\pi_{\theta_i}} [ \nabla_{\theta_i} \log \pi_{\theta_i}(\mathbf{a}|\mathbf{o}) Q_\phi^\pi(\mathbf{o}, \mathbf{a}) ].
    \label{eq1}
\end{equation}

\noindent \textbf{Distributional reinforcement learning.} Unlike traditional RL whose target is to maximize the expected total return, distributional RL \cite{bellemare2017distributional} explicitly considers the randomness of the return distribution. 
The expected discounted return $Q^{\pi}$ can be written as:
\begin{equation}
    Q^{\pi}(s, a) = \mathbb{E}[Z^{\pi}(s, a)] = \mathbb{E}[\sum_{t=0}^{\infty}\gamma^tr(s_t, a_t)],
    \label{eq2}
\end{equation}
where $s_t \sim P(\cdot|s_{t-1},a_{t-1}), a_t \sim \pi(\cdot|s_t), s_0= s, a_0 =a$. $Z^{\pi}(s, a)$ is the return distribution, covering all sources of intrinsic randomness, including reward function, state transition, stochastic policy sampling, and systematic stochasticity.
In the distributional RL, we directly model the random return $Z^{\pi}(s,a)$ instead of its expectation. The distributional Bellman equation can be defined as: 
\begin{equation}
    Z(s,a) \overset{D}{=} r(s,a) + \gamma P^{\pi}Z(s,a),
    \label{eq3}
\end{equation}
where $\overset{D}{=}$ means the two sides of the equation are distributed according to the same law,
$P^{\pi}Z(s,a)\overset{D}{=} Z(S',A')$ and $S' \sim P(\cdot|s,a)$, $A' \sim \pi(\cdot|S')$.

{In general, the distribution of the random variable $Z(s, a)$ is represented} as $\frac{1}{K}\sum_{j=1}^{K}{\delta_{G_{\phi}^j(\mathbf{o}, \mathbf{a})}}$ where $\delta_{G_{\phi}^j(\mathbf{o}, \mathbf{a})}$ denotes a Dirac with respect to the support $G_{\phi}^j(\mathbf{o}, \mathbf{a}) \in \mathbb{R}$. Furthermore, each $G_{\phi}^j(\mathbf{o}, \mathbf{a})$ can be estimated through optimizing the quantile Huber loss \cite{dabney2018implicit}: 

\begin{equation}
    \mathcal{L}_{QR} = \frac{1}{K^2}\sum_{j'=1}^K\sum_{j=1}^K [\rho_{\omega_j}(\delta_{ij})],
\label{eq10}
\end{equation}
where $\rho_{\omega_j}(\mu)= |\omega_j-\mathbbm{1}_{\mu \leq 0}|\mathcal{L}_{\kappa}(\mu)$ and $\delta_{ij} =r+\gamma G_j(\mathbf{o}', \mathbf{a}')- G_j(\mathbf{o}, \mathbf{a})$.
$\mathbbm{1}_{\mu \leq 0}$ is an indicator function which is $1$ 
when $\mu\leq 0$ and $0$ otherwise.
$\kappa$ is a pre-fixed threshold and the Huber loss $\mathcal{L}_{\kappa}(\mu)$ is given by,

\begin{equation}
    \mathcal{L}_{\kappa}(\mu)= \frac{1}{2}\mu^2\mathbbm{1}_{[|\mu| \leq \kappa]} + \kappa(|\mu|-\frac{1}{2}\kappa)\mathbbm{1}_{[|\mu| > \kappa]}.
    \label{eq11a}
\end{equation}

\begin{table}\centering
\caption{{Notations of algorithm}}
  \begin{tabular}{c|p{15em}}
    \hline
    \textbf{Notation} & \textbf{Meaning } \\
    \hline
    $p$ & The numbers of opponents \\
    $\mu_{\psi_{ik}}$ & Predicted opponent model for opponent $k$ of agent $i$  \\
    $o_i$ & Local observation of agent $i$\\
    $\hat{a}_{ik}^t$ & Predicted opponent actions for opponent $k$ of agent $i$ \\
    $\pi_{\theta_i}$ & Policy network of agent $i$\\
    $a_i^t$ & Action of agent $i$ \\
    $\rho_{\theta_i,\psi_i}$ & Final agent policy distribution\\
    $Z_i(\mathbf{o}^t,\mathbf{a}^t)$   & Return distribution of agent $i$\\
    $G_{\phi_i}$ & CDC network\\
    $\omega$ & Quantiles\\
    $\delta_z$ & Dirac at $z \in \mathbb{R}$\\ 
    $K$ & Quantile number\\
    $\alpha_i$ & hyperparameter of exploration\\
    \hline
  \end{tabular}
  \label{T_notafication}
\end{table}

\section{Methodology}\label{s4}
\subsection{The Overall Framework}
\begin{figure}[!t]
\vspace{-5pt}
    \centering
    \includegraphics[width=0.48\textwidth]{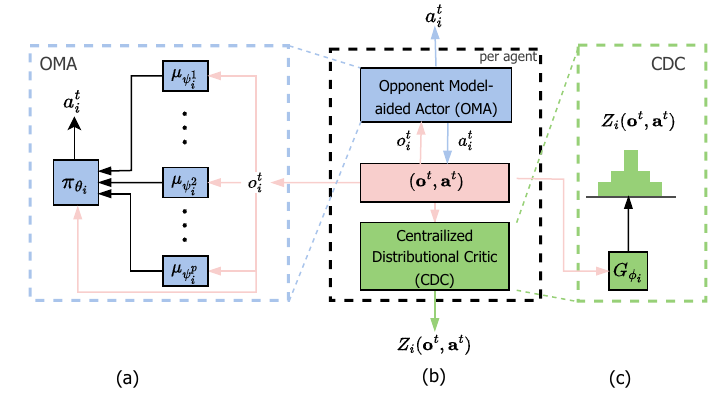}
    \caption{{The DOMAC framework.  In OMA, the speculative opponent model $\mu_i$ takes the local observation $o^t_i$ as input and outputs the prediction of the opponent behaviours. The agent policy network takes action by considering the predicted opponent's actions. The CDC takes the joint observation $\mathbf{o}^t$ and action $\mathbf{a}^t$ of the controlled agents into the agent critic network $G_{\phi_i}$ and outputs the return distribution $Z_i(\mathbf{o}^t,\mathbf{a}^t)$.} }
    \label{DOMAC}
\end{figure}

Our training follows the CTDE setting, i.e., we have the access to the observations and actions of the team we control.
Note that, however, we \textbf{do not know} the observations and actions of the opponent team during the training. Figure \ref{DOMAC} depicts the overall framework of the proposed \textit{distributional opponent model aided actor-critic} (DOMAC) algorithm. {We show the meaning of the notifications in Table \ref{T_notafication}.}

{Within the DOMAC framework, each controlled agent $i$ incorporates two primary constituents: the Opponent Model Aided Actor (OMA) and the Centralized Distributional Critic (CDC). For $p$ opponents, agent $i$ employs $p$ individualized conjectural opponent models, wherein each model is instantiated as a distinct neural network. These models process the agent's localized observational data and sample actions at each timestamp to ascertain the presumptive action distributions of the opponents. The OMA leverages these models to infer potential joint actions, representing the agent's conjectures concerning the opponents' prospective moves. The agent $i$'s own action distribution is subsequently determined as a composite function of the outputs from the OMA and the observations, effectively assimilating the agent's tactical considerations with the inferences drawn from the opponent models by performing a weighted aggregation. To enhance predictive precision, the agent samples multiple opponents' joint actions. Intriguingly, the opponent models' training regimen is predicated on the rewards obtained by the agent's policy alone, eschewing reliance on the opponents' genuine actions and underscoring outcome-oriented feedback loops. Meanwhile, the CDC's role is to evaluate the return distribution associated with the agent's policy, preferring a distributional perspective over singular expected value estimations. This methodological choice for estimating returns confers a more nuanced depiction, which polishes the refinement process for the agent's strategic policy.}

Note that each agent in the controlled team has an independent set of OMA and CDC, and the only shared are their observations and actions. This is critical for extracting the individual knowledge of each agent, which may be heterogeneous for their different observations, current statuses, etc. {To speed up the learning, we share the parameters of the agent networks across all controlled agents but a one-hot encoding of the $agent_{id}$ is concatenated onto each agent's observations to induce heterogeneities of the inputs. In scenarios where the agents within a system are homogeneous, both functionally and behaviorally, the utilization of a shared
OMA network is feasible for these homogeneous agents.} %
  Generally, the rationales of DOMAC can be explained as the same closed optimization loop as that of the original actor-critic framework. That is, OMA receives the local information with the beliefs of modelled opponents (i.e., predicted opponents' actions). It improves decision-making quality by adjusting opponent models and actively selecting the actions with higher evaluations according to CDC, while CDC, in turn, evaluates the improved policy. %
In the following, we first show that optimizing the OMA can be formulated as a policy gradient theorem. Then we formulate the optimization of CDC by using a quantile regression loss. Finally, we transform the theoretical findings into a practical algorithm.

\begin{figure*}[!t]
    \centering
    \includegraphics[width=0.85\textwidth]{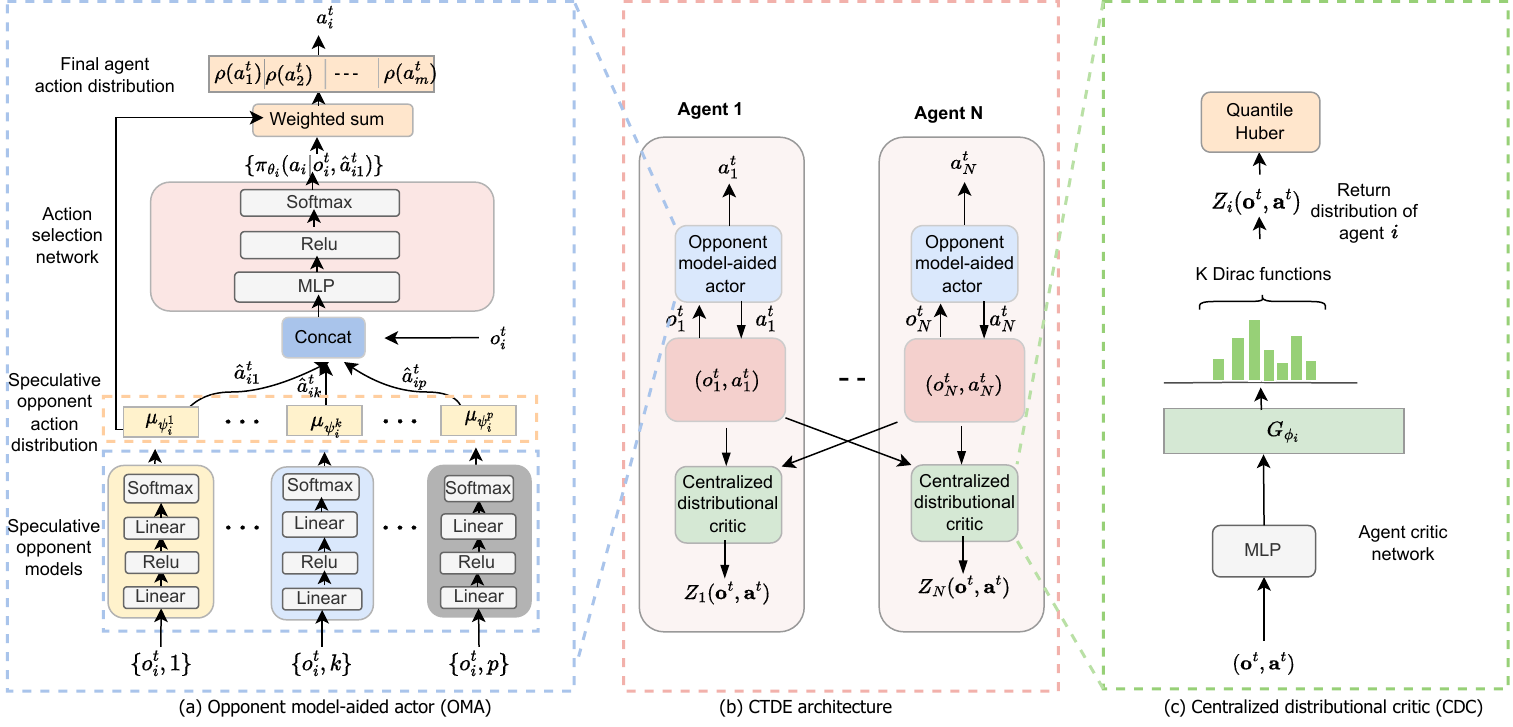}
    \vspace{-5pt}
    \caption{{An illustration of our DOMAC network architecture. (a) Opponent model-aided actor: Each model contains $p$ speculative opponent models, {which take the local observation $o^t_i$ and opponent index $k$} as inputs {to} {predict the opponents' actions}. Then, the {action selection network} $\pi_{\theta_i}$ takes the joint predicted action $\{\hat{a}^t_i\}$ together with the $o^t_i$ and $a^{t-1}_i$ as input, and outputs a distribution over the agent $i$'s own {actions, which is weighed according to the probabilities of predicted opponents' actions $\mu^k_{\phi_i}$ for $1 \leq k \leq p$}.  (b) CTDE architecture: we have the access to the observations and actions of the team we control. (c) Centralized distributional critic: the agent critic network takes the joint observation $\mathbf{o}^t$ and action $\mathbf{a}^t$ of the controlled agents and outputs $G_{\Phi_i}$ as the approximation of return distribution.}}
    \label{DOMAC_new}
\end{figure*}
\subsection{Opponent Model Aided Actor} \label{OMA}

{Suppose the agent $i$ has total $p$ opponents in the game, it {represents} them by $p$ {independent} speculative opponent models. Specifically in Figure~\ref{DOMAC_new}(a), the opponent model for opponent $k$ of agent $i$ is given as $\mu_{\psi_{ik}}$ with trainable parameter $\psi_{ik}$.
Once the agent $i$ receives a local observation $o^t_i$ at time step $t$, each of its opponent models takes $o^t_i$ and the opponent index {k} as input, and outputs a distribution over which the opponent $k$'s actions $\hat{a}^t_{ik}$ can be predicted.}
 
{Then, the agent policy $\pi_{\theta_i}$ takes the joint predicted action $\hat{a}^t_i = [\hat{a}^t_{i1}, \dots, \hat{a}^t_{ip}]$ {of all opponents} {with $o^t_i$ as input}, and outputs a distribution over the agent $i$'s action space $A_i$. 
The final action probability distribution of the agent is the sum of the actor outputs weighted by the probability of corresponding opponent joint action {(Equation~\ref{eq4})}. To increase the accuracy, agent $i$ samples and aggregate multiple opponents' joint actions.}

{Then, {we connect our OMA module with the Policy Gradient Theorem \cite{sutton2000policy}.}} For clarity, we omit time step $t$ in all formulas below.
Let $\mu_{\psi_{ik}}$ and $\pi_{\theta_i}$ be parameterized with the trainable parameters $\psi_{ik}$ and $\theta_i$, respectively, and $\psi_i = \{ \psi_{ik} \}$ be the set of parameters of all speculative opponent models maintained by the agent $i$.
We assume that these opponent models are independent with each other.
Given an observation $o_i$, the agent $i$'s OMA calculates its action probability distribution as:
\begin{align}
     \rho_{\theta_i, \psi_i}(a_i | o_i) &= \sum_{\hat{a}_i} [ \pi_{\theta_i}(a_i | \hat{a}_i, o_i)  P(\hat{a}_{i} | o_i)] \notag\\
     &= \sum_{\hat{a}_i} [ \pi_{\theta_i}(a_i | \hat{a}_i, o_i) \prod_{k=1}^{p} \mu_{\psi_{ik}}(\hat{a}_{ik} | o_i) ],
    \label{eq4}
\end{align}
where $\hat{a}_i = [\hat{a}_{i1}, \dots, \hat{a}_{ip}]$ is an opponent joint action predicted by the agent $i$'s opponent models. 
Then, the objective of agent $i$ is to maximize its total expected return, which is determined by the parameters $\theta_i$ and $\psi_i$:
\begin{equation}
\begin{aligned}
     J(\theta_i, \psi_i) = \mathbb{E}_{\mathbf{o}\sim\tau, a_i\sim\rho_{\theta_i, \psi_i}} [ \mathbb{E}(Z_i(\mathbf{o}, \mathbf{a}))],
     \label{eq5}
\end{aligned}
\end{equation}
where $(\mathbf{o}, \mathbf{a})$ represents the joint observation and joint action of the agent $i$'s team, and $\tau$ represents the trajectory distribution induced by all agents' policies.
Additionally, we add the entropy of the policy into the objective function (\ref{eq5}) to ensure an adequate exploration, which is proved to be effective for improving the agent performance \cite{mnih2016asynchronous}. Thus, the OMA's final objective can be defined as: 
\begin{equation}
\begin{aligned}
     J(\theta_i, \psi_i) = \mathbb{E}_{\mathbf{o}, a_i} [ \mathbb{E}( Z_i(\mathbf{o}, \mathbf{a})) - \alpha_i \log \rho_{\theta_i, \psi_i}(a_i|o_i)],
     \label{eq6}
\end{aligned}
\end{equation}
where $\alpha_i$ is a hyperparameter controlling the degree of exploration.
Next, we mathematically derive the closed form of the policy gradient and opponent model gradient respectively, which is summarized in the following proposition. 

\begin{theorem}\label{prop1}
In a POMG, under the opponent {modelling} framework defined, the gradient of parameters $\theta_i$ for the policy of the agent $i$ is given as:
\begin{equation}
\begin{split}
    \nabla_{\theta_i}J(\theta_i,\psi_i) &= \mathbb{E}_{\rho_{\theta_i,\psi_i}}\Bigl[\nabla_{\theta_i} \log (\rho_{\theta_i,\psi_i}) \cdot \\
    & [\mathbb{E}(Z_i) - \alpha_i \log (\rho_{\theta_i,\psi_i}) -\alpha_i]\Bigr],
\end{split}
\label{eq7}
\end{equation}
and the gradient of the parameters $\psi_{ik}$ for the opponent model $\mu_{\psi_{ik}}$ is given as:
\begin{equation}
\begin{split}
    \nabla_{\psi_{ik}}J(\theta_i,\psi_i) &= \mathbb{E}_{\rho_{\theta_i,\psi_i}}\Bigl[\nabla_{\psi_{ik}}\log(\rho_{\theta_i,\psi_i}) \cdot  \\
    &[\mathbb{E}(Z_i)- \alpha_i \log (\rho_{\theta_i,\psi_i}) - \alpha_i]\Bigr],
\end{split}
\label{eq8}
\end{equation}
where $\rho_{\theta_i, \psi_i}$ is defined in Equation (\ref{eq4}).
\end{theorem}
\noindent \textit{Proof.} 
For notational convenience, $Z_i(\mathbf{o},\mathbf{a})$ can be considered as $Z_i$, and $\rho_{\theta_i,\psi_i}(a_i|o_i)$ can be simplified as  $\rho_{\theta_i,\psi_i}$. Then  
the objective in Equation (6) can be written as:

\begin{equation}
    J(\theta_i, \psi_i) = \sum_{a_i}\rho_{\theta_i,\psi_i}[\mathbb{E}(Z_i))-\alpha_i \log \rho_{\theta_i,\psi_i}].
\end{equation}
The policy gradient of parameter $\theta_i$ can be calculated as:
\begin{equation}
\begin{split}
    \nabla_{\theta_i}J(\theta_i,\psi_i) &= \sum_{a_i} \nabla_{\theta_i} \rho_{\theta_i,\psi_i}\mathbb{E}(Z_i) \\
    &-\alpha_i \sum_{a_i} \nabla_{\theta_i}[\rho_{\theta_i,\psi_i}\log \rho_{\theta_i,\psi_i}]\\
    & = \sum_{a_i}\nabla_{\theta_i} \rho_{\theta_i,\psi_i}\mathbb{E}(Z_i)\\
    &- \alpha_i \sum_{a_i} \nabla_{\theta_i}
    \rho_{\theta_i,\psi_i}[\log \rho_{\theta_i,\psi_i}+1]\\
    & = \sum_{a_i}\nabla_{\theta_i}\rho_{\theta_i,\psi_i}[\mathbb{E}(Z_i)-\alpha_i \log \rho_{\theta_i,\psi_i}-\alpha_i ]\\
    & =\sum_{a_i} \rho_{\theta_i,\psi_i}\nabla_{\theta_i}\log \rho_{\theta_i,\psi_i}[\mathbb{E}(Z_i) \\
    &-\alpha_i \log \rho_{\theta_i,\psi_i} -\alpha_i]\\
    & = \mathbb{E}_{\rho_{\theta_i,\psi_i}}\Bigl[\nabla_{\theta_i}\log\rho_{\theta_i,\psi_i}[\mathbb{E}(Z_i)\\
    &-\alpha_i \log \rho_{\theta_i,\psi_i} -\alpha_i]\Bigr].
\end{split}
\end{equation}
Since $\psi_i = \{\psi_{ik}\}$, then 
$\nabla_{\psi_i}J(\theta_i, \psi_i)= \{\nabla_{\psi_{ik}} J(\theta_i, \psi_i)\}$.
For each opponent agent $k$, the gradient of parameter $\psi_{ik}$ can be calculated as:
\begin{equation}\label{EqA.2}
\begin{split}
    \nabla_{\psi_{ik}}J(\theta_i,\psi_i) &= \sum_{a_i} \nabla_{\psi_{ik}} \rho_{\theta_i,\psi_i}\mathbb{E}(Z_i)\\
    &-\alpha_i \sum_{a_i} \nabla_{\psi_{ik}}[\rho_{\theta_i,\psi_i}\log \rho_{\theta_i,\psi_i}].
\end{split}
\end{equation} 
Based on the Equation (4), we can conclude that the parameters of $\theta_i$ and ${\psi_{ik}}$ are independent. Then we can further obtain that
\begin{equation}
    \begin{split}
    \nabla_{\psi_{ik}}J(\theta_i,\psi_i) &=\mathbb{E}_{\rho_{\theta_i,\psi_i}}\Bigl[\nabla_{\psi_{ik}}\log\rho_{\theta_i,\psi_i}[\mathbb{E}(Z_i)\\ &-\alpha_i \log \rho_{\theta_i,\psi_i}-\alpha_i]\Bigr].
    \end{split}
\end{equation}\label{EqA.3} 

Note that $\rho_{\theta_i, \psi_i}$ conditions on $\psi_{ik}$ as shown in Equation (\ref{eq4}). Therefore, the operation $\nabla_{\psi_{ik}}\log(\rho_{\theta_i,\psi_i})$ in Equation (\ref{eq8}) enables the gradients to naturally backpropagate to the opponent models.
This proposition states that OMA can adjust the parameters $\theta_i$ and $\psi_i$ towards maximizing the objective by following the gradient. 
Intuitively speaking, the OMA directly searches the optimal policy and reliable opponent models by looking at the agent's objective, which does not require access to opponents' true information.
Although it may look weak to use the objective as the only learning signal, we will show that the gradient of the objective is already sufficient for training reliable opponent models and thereby good actors.

\subsection{Centralized Distributional Critic} \label{CDC}

To enhance training and provide a more delicate evaluation of the OMA, we adopt a distributional perspective to evaluate agents' performance by {modelling} their return distributions \cite{bellemare2017distributional}. 
We propose to learn a centralized distributional critic (CDC) for each controlled agent $i$.  {After all agents in the controlled team sample their actions, each controlled agent $i$'s CDC network $G_{\phi_i}$ takes as input the joint observation $\mathbf{o}^t$ and joint action $\mathbf{a}^t$ of its team, and
computes a return distribution $Z_i(\mathbf{o}^t, \mathbf{a}^t)$.}
Following \cite{dabney2018distributional}, we train the CDC with the quantile regression technique.
Specifically, we implement the CDC for the agent $i$ as a deep neural network $G_{\phi_i}$ with $\phi_i$ being the trainable parameters. Given the joint observation and joint action $(\mathbf{o}, \mathbf{a})$ of the agent $i$'s team, $G_{\phi_i}$ outputs a $K$-dimensional vector $\{G_{\phi_i}^j(\mathbf{o},\mathbf{a})\}_{j=1,\cdots,K}$. The elements of this output are the samples that approximate the agent $i$'s return distribution $Z_i(\mathbf{o}, \mathbf{a})$. That is, we can represent the distribution approximation as $\frac{1}{K}\sum_{j=1}^{K}{\delta_{G_{\phi_i}^j(\mathbf{o}, \mathbf{a})}}$ where $\delta_z$ denotes a Dirac at $z \in \mathbb{R}$. An ideal approximation can map these $K$ samples to $K$ fixed quantiles $\{\omega_j=\frac{j}{K}\}_{ j=1,\cdots,K}$ so that $P(Z_i(\mathbf{o}, \mathbf{a})\leq G_{\phi_i}^j(\mathbf{o}, \mathbf{a}))= \omega_j$. 
To approach that, we compute the target distribution approximation $\hat{G}_{\phi_i}(\mathbf{o}, \mathbf{a})$ based on the distributional Bellman operator,
\begin{equation}
    \hat{G}_{\phi_i}(\mathbf{o}, \mathbf{a}) = \{r(s, \mathbf{a})+\gamma G_{\phi_i}^j(\mathbf{o}', \mathbf{a}^*) \}_{j=1,\cdots,K} ,
\label{eq9}
\end{equation}
where $\mathbf{o}'\sim \tau, \mathbf{a}^* =  \arg \max_{\mathbf{a}'}\mathbb{E}[G_{\phi_i}(\mathbf{o}', \mathbf{a}')].$
The objective of the CDC training then becomes minimizing the distance between $\hat{G}_{\phi_i}$ and $G_{\phi_i}$. To this end, we use the
Quantile Huber loss \cite{dabney2018distributional}, 
\begin{equation}
    \mathcal{L}_{QR} = \frac{1}{K^2}\sum_{j'=1}^K\sum_{j=1}^K [\rho_{\omega_j}(\hat{G}_{\phi_i}^{j'}(\mathbf{o},\mathbf{a}) - G_{\phi}^j(\mathbf{o},\mathbf{a}))],
\label{eq10a}
\end{equation}
where $\rho_{\omega_j}(\mu)= |\omega_j-\mathbbm{1}_{\mu \leq 0}|\mathcal{L}_{\kappa}(\mu)$.
$\mathbbm{1}_{\mu \leq 0}$ is an indicator function which is $1$ 
when $\mu\leq 0$ and $0$ otherwise.
$\kappa$ is a pre-fixed threshold and the Huber loss $\mathcal{L}_{\kappa}(\mu)$ is given by,
\begin{equation}
    \mathcal{L}_{\kappa}(\mu)= \frac{1}{2}\mu^2\mathbbm{1}_{[|\mu| \leq \kappa]} + \kappa(|\mu|-\frac{1}{2}\kappa)\mathbbm{1}_{[|\mu| > \kappa]}.
    \label{eq11}
\end{equation}

\subsection{Practical Algorithm}
We now present a tangible algorithm for training OMA and CDC. First, we describe a sampling trick that is essential for training. {{For OMA, the controlled agent maintains a set of opponent models, each of which is instantiated by an independent neural network and corresponds to an opponent.} Thus, the number of opponent models and overall network size is linear with the number of opponents, which can be easily handled by GPU parallel training. However, 
computing the marginal distribution $\rho_{\theta_i, \psi_i}$ can be exponentially costly concerning the dimensionality of the opponents' action space according to Equation (\ref{eq4}).} Formally, each agent $i$'s opponent $k$ has 
$|A_{ik}|$ actions. Then calculating the exact $\rho_{\theta_i, \psi_i}$ requires traversing $|A_{i1}| \times \cdots \times |A_{ip}|$ opponent joint  actions, which quickly becomes intractable as $p$ increases. {To address this issue}, in our algorithm, we apply a sampling trick that samples a set of actions 
$\hat{a}_{ik}= (\hat{a}_{ik1},\cdots, \hat{a}_{ikl})$
from the output of the speculative opponent model $\mu_{\psi_{ik}}$ 
for each opponent $k$, where $l$ controls the size of sampled actions. 
Then, $\rho_{\theta_i, \psi_i}(a_i|o_i)$ can be approximated as:
\begin{equation}
    \Bar{\rho}_{\theta_i, \psi_i}(a_i|o_i) = \sum_l \pi_{\theta_{i}}(a_i|o_i, \{\hat{a}_{ikl}\})\prod_{k=1}^p \mu_{\psi_{ik}}(\hat{a}_{ikl}|o_i).\notag
    \label{eq12}
\end{equation}
The action $a_i$ can be sampled from $\Bar{\rho}_{\theta_i, \psi_i}(a_i|o_i)$ for each agent $i$. We use empirical samples $\{G_{\phi_i}^j(\mathbf{o},\mathbf{a})\}_{j=1,\cdots,K}$ to approximate $Z_i(\mathbf{o}, \mathbf{a})$. The objective function of OMA is,
\begin{equation}
\begin{aligned}
     J(\theta_i, \psi_i) = \mathbb{E}_{\mathbf{o}, a_i} [ \frac{1}{K}\sum_{j=1}^K G_{\phi_i}^j(\mathbf{o},\mathbf{a}) - \alpha_i \cdot \log {\Bar{\rho}}_{\theta_i, \psi_i}(a_i|o_i)].
     \label{eq13}
\end{aligned}
\end{equation}
We incorporate the proposed OMA and CDC into the on-policy framework. In general, the training procedure of DOMAC is similar to that of other on-policy multi-agent actor-critic algorithms \cite{foerster2018counterfactual}, where the training data is generated on the fly. At each training iteration, the algorithm first generates training data with the current OMA and CDC. Then, we compute the objectives of OMA and CDC from the generated data, respectively. At last, the optimizer updates the parameters $\phi_i$, $\theta_i$, and $\psi_i$ accordingly. {After that}, the updated OMA and CDC are used in the next training iteration. In practice, we can parallelize the training, which is a common technique to reduce the training time \cite{iqbal2019actor}. In such cases, the training data is collected from all parallel environments, and actions are sampled and executed in respective environments concurrently. We summarize the training procedure in Algorithm \ref{alg:algorithm}.

\begin{algorithm}[tb]
\footnotesize
\caption{Training Procedure of DOMAC}
\label{alg:algorithm}
\textbf{Require:} A POMG environment \textit{env}, a back-propagation optimizer $\texttt{Opt}$, number of episodes $E$, number of forward steps $T_f$, and $\pi_{\theta_i}$, $\mu_{\psi_i}$, $G_{\phi_i}$ with initialized parameters $\theta_i$, $\psi_i$, and $\phi_i$ for each agent $i$.
\begin{algorithmic}[1]
\For{$e$ =1, \dots, $E$}
\State $t \leftarrow 1$; Reset \textit{env} to obtain initial observations $\mathbf{o}^1$
\While{\textit{env} is not done} \texttt{// Generate data}
\For{$t^* = t,\dots, t + T_f -1$ }
\State Sample each agent $i$'s action $a_i^{t^*} \sim \rho_{\theta_i, \psi_i}(a_i^{t^*}|o_i^{t^*})$
\State Form the controlled team's joint action $\mathbf{a}^{t^*}=\{a_i^{t^*}\}$ 
\State Execute $\mathbf{a}^{t^*}$ to obtain $\mathbf{r}^{t^*}$ and $\mathbf{o}^{t^*+1}$
\State Store transition data $(\mathbf{o}^{t^*},\mathbf{a}^{t^*}, \mathbf{r}^{t^*}, \mathbf{o}^{t^*+1})$
\EndFor
\For{$1 \leq i \leq n$} \texttt{// Update OMA, CDC}
\State Calculate $Z_i$, $\Bar{\rho}_{\theta_i, \psi_i}$ with collected data
\State Updates $G_{\phi_i}$ by minimizing Equation (\ref{eq10})
\State Updates $\pi_{\theta_i}$ by minimizing Equation (\ref{eq13})
\State Updates $\mu_{\psi_i}$ by minimizing Equation (\ref{eq13})
\EndFor
\State $t \leftarrow t + T_f$
\EndWhile
\EndFor
\end{algorithmic}
\end{algorithm}

\section{Experiments}\label{s5}
{To thoroughly assess the DOMAC algorithm, we evaluate our model in three {widely-used} partial observable multi-agent environments {for various settings}. {The environments are} the predator-prey game based on the Multi-agent Particle Environment \cite{lowe2017multi}, the {Pommerman} Environment \cite{DBLP:journals/corr/abs-1809-07124} and {the} StarCraft Multiagent Challenge (SMAC) {suit} \cite{samvelyan2019starcraft}}. 
{{From the experiments, we} aim to answer the following questions:} {(1): Does the DOMAC yield superior performance than {SOTA} baseline methods (Figure \ref{fig5}, \ref{fig6} and Table \ref{T1})?} {(2): Are the main components of DOMAC, i.e.,  the OMA and CDC, necessary {and effective} (Figure \ref{fig7}, Figure \ref{fig8}, Figure \ref{fig9}, and Figure \ref{fig12})?} {(3): How do the key hyperparameters in DOMAC affect the final results (Figure \ref{fig10}, Figure \ref{fig11} and Table \ref{T2})?}


\subsection{The Environment Setup} \label{MA environments}

\noindent \textbf{Setup for the predator-prey environment.} 

\begin{figure}[!t]
\footnotesize
    \centering
    \subfigure[]{\includegraphics[width=.20\textwidth]{{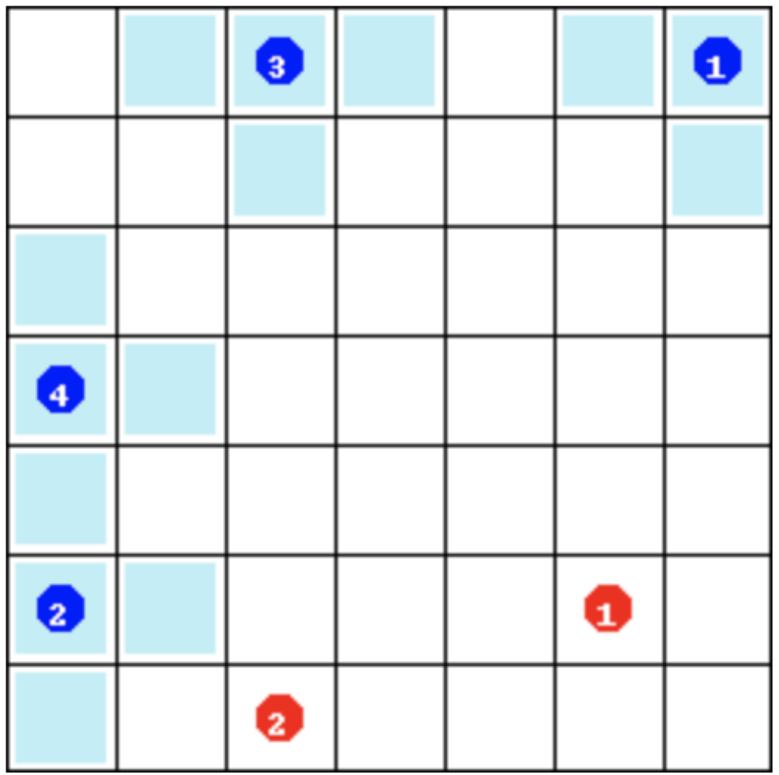}}\label{predator_prey}}
    \subfigure[]{{\includegraphics[width=.226\textwidth]{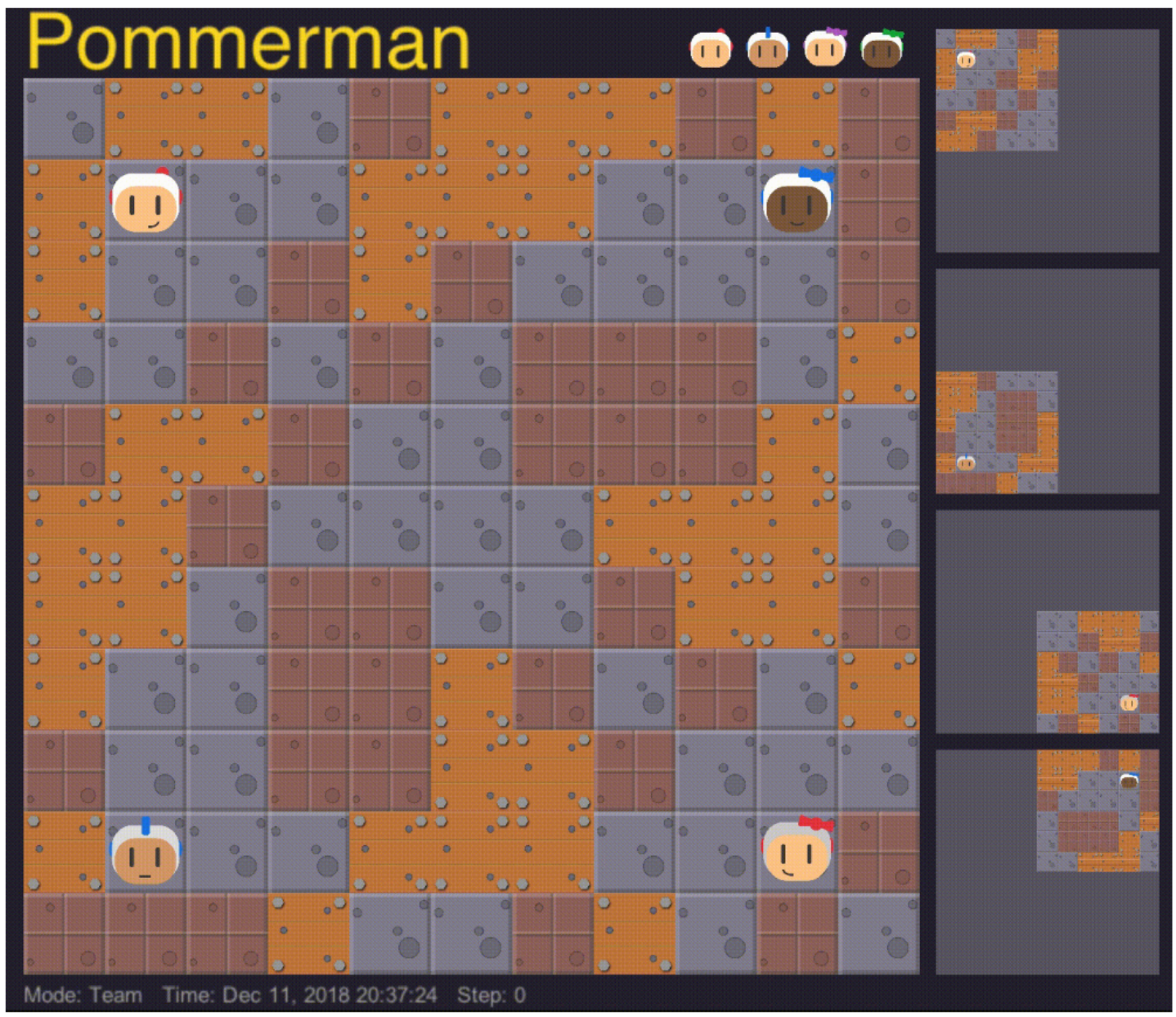}}\label{pommerman}}
    
    \caption{\textbf{State visualization of benchmark environments.} (a) The state of a \texttt{PP-3v1} game, where blue vertices and red vertices denote the predators and prey respectively. (b) The image-based state for the pommerman environment.}
    \label{fig2}
\end{figure}

\noindent In the predator-prey environment (Figure \ref{predator_prey}), the player controls multiple cooperating predators to catch faster-moving preys in 500 iterations. Every prey has a health of 10. A predator moving within a given range of the prey lowers the prey's health by 1 point per time step. Lowering the prey's health to 0 can kill the prey. If there is at least one prey surviving after 500 iterations, the prey team wins. All agents choose among five moving actions. $L$ landmarks (gray circles) that impede the agents' way are randomly placed in the environment at the start of the game. 
Each predator obtains the relative positions and velocities of the agents, and the positions of the landmarks as an observation. 
 We evaluate DOMAC in two scenarios. In one scenario, there exist three predators and one prey and we denote it as \texttt{PP-3v1}. Another one includes five predators and two preys and we denote it as \texttt{PP-5v2}. The number of landmarks is $2$ in both settings. 

\noindent \textbf{Setup for the pommerman environment.} \\
\noindent This environment involves four agents. In every step, these agents can move in one of four directions, place a bomb, or do nothing. As shown in Figure \ref{pommerman}, 
{agents get local observations within their field of view $5\times 5$, which contains information (board, position, ammo) about the map}. An empty grid allows any agents to enter it. A wooden grid cannot be entered but can be destroyed by a bomb. A rigid grid is unbreakable and impassable. When a bomb is placed in a grid, it will explode after 10 time steps. The explosion will destroy any wooden grids and agents which locate within 4-grids away from the bomb.
If all agents of one team die, the team loses the game. The game will be terminated after $1000$ steps no matter whether there is a winning team or not. Agents get $+1$ reward if their team wins and $-1$ reward otherwise.
The details of the environments are provided in the supplementary material. 
We carry out the experiments in two different settings. In the first setting, four agents fight against each other, denoted as \texttt{Pomm-FFA}. The second setting is a two-team competition where each team has two agents, denoted as \texttt{Pomm-Team}.

\noindent {\textbf{Setup for StarCraft environments}}

\begin{figure}[!t]
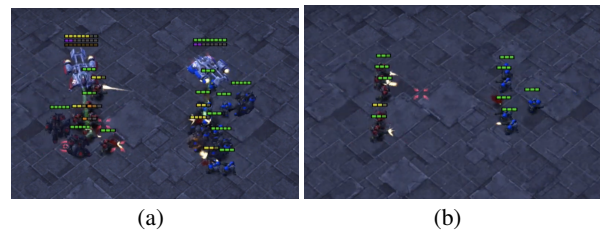

\footnotesize
    \centering
    \subfigure[]{\includegraphics[width=.215\textwidth]{{SMAC-mmm2.pdf}}\label{sc2_mmm2}}
    \subfigure[]{{\includegraphics[width=.22\textwidth]{SMAC-5m6m.pdf}}\label{sc2_2s3z}}
  
    \caption{\textbf{State visualization of StarCraft II,
} (a) The state of a \texttt{MMM2} game,  (b) The state of a \texttt{5m\_vs\_6m} game.}
    \label{fig4}
\end{figure}

\noindent {{For the} SMAC benchmark \cite{samvelyan2019starcraft}, {we adopt the v4.10 StarCraft II simulator}. In SMAC, {each agent is represented by a troop unit with different attacking abilities}. {The agents} can move in four directions and are allowed to perform the attack action only if the enemy is within its shooting range. 
{As depicted in Figure \ref{fig4}, each agent obtains local observations within its visual radius, capturing details like distance, relative location, health, shield, and unit type for all units (friendly and hostile) within a defined circular sector of the map.} At each time step $t$, {each} agent receives {the same joint} reward equal to the {cumulative} damage dealt to the enemy {team until $t$}. In addition, {all agents receive} a bonus of 10 points after {an} opponent {is killed} and 200 points after {wiping all enemy team}. {In addition}, these rewards are all normalized to ensure the maximum cumulative reward achievable in an episode is 20 points. We evaluate our method in four {widely-used} scenarios: \texttt{2s3z} (easy and symmetric heterogeneous), \texttt{1c3s5z} (easy and symmetric heterogeneous), \texttt{5m\_vs\_6m} (hard and asymmetric homogeneous), and \texttt{MMM2} (super hard and symmetric heterogeneous)}. We provide the details of the three environments in Appendix \ref{Appendix-B}.
 

\subsection{Baselines and Algorithm Configuration.}\label{section5_2}
\noindent \textbf{Baselines.}\\
To the best of our knowledge, this is the first work that considers the opponent {modelling} problem with purely local information. Because the existing works on opponent {modelling} all require access to opponents' true information, they cannot work in our setting. Therefore, we compare the proposed DOMAC algorithm with one of the most popular multi-agent RL algorithms: multi-agent actor-critic (MAAC) \cite{lowe2017multi} which is the discrete version of multi-agent deep deterministic policy gradient (MADDPG). Because MAAC does not incorporate opponent {modelling}, it is compatible with our setting where opponents' true information is not available. Also, we further integrate MAAC with our OMA and CDC respectively to generate another two baselines. Let OMAC denote the baseline that combines MAAC with OMA and {DMAC} denote the baseline that combines MAAC with CDC. The comparison with OMAC and DMAC can demonstrate the impact of OMA and CDC, and thereby justify our design. {To comprehensively verify the performance of DOMAC, we also compare it with a state-of-the-art (SOTA) policy gradient algorithm, MAPPO, which has achieved satisfactory performance in many environments \cite{yu2022surprising}.} Furthermore, we evaluate an upper bound setting (denoted as UB) where we directly use the actual opponent policy to predict the opponents' actions during training. Because the actual opponent policy can provide accurate opponent information to the agent actors, which is the learning goal of our speculative opponent models, the performance of UB can be regarded as the upper bound of the performance of DOMAC.

\begin{figure}[htp]
\vspace{-1.0em}
   \centering
    \subfigure[ \texttt{PP-3v1}. ]{\includegraphics[width=.22\textwidth]{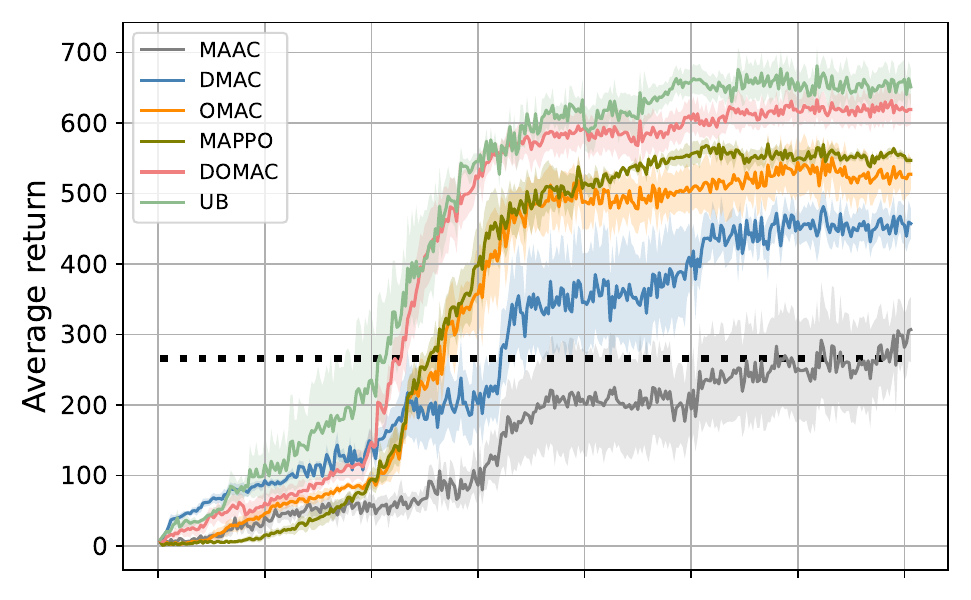}\label{5:a}} \vspace{-1mm}
    \subfigure[\texttt{Pomm-FFA}.]{\includegraphics[width=.22\textwidth]{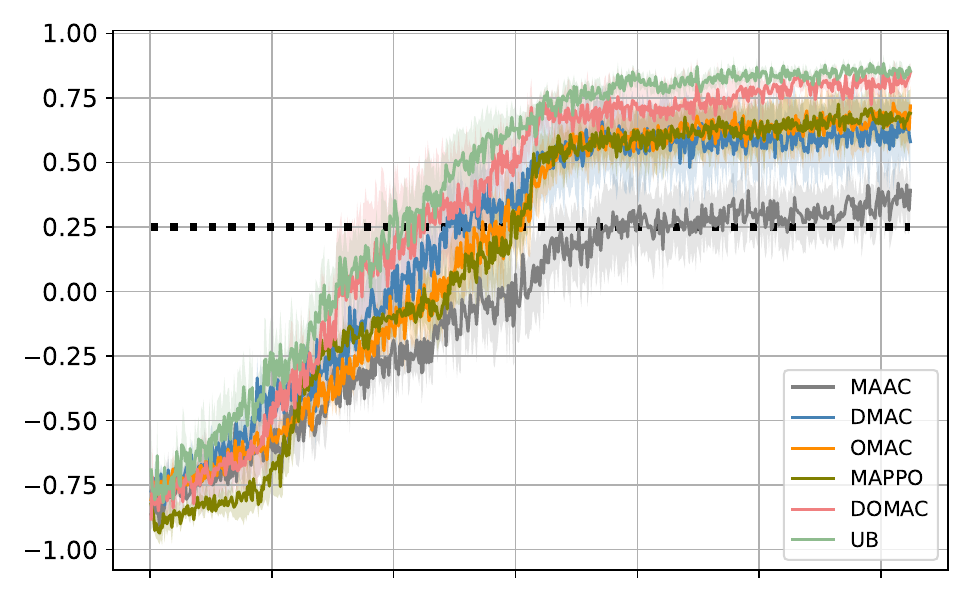}\label{5:b}} \vspace{-1mm}
    \subfigure[ \texttt{PP-5v2}.]{\includegraphics[width=.22\textwidth]{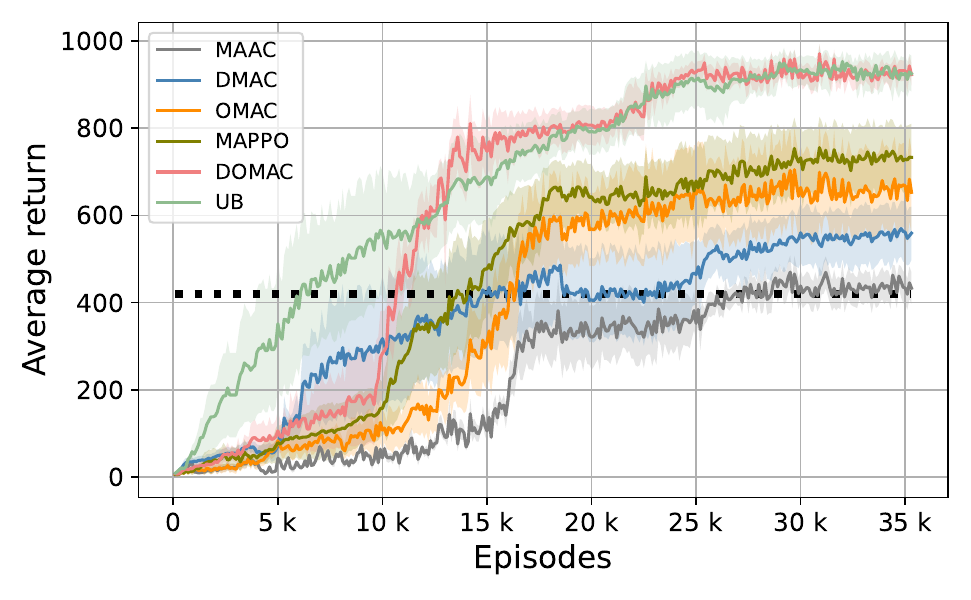}\label{5:c}}
    \subfigure[ \texttt{Pomm-Team}.]{\includegraphics[width=.22\textwidth]{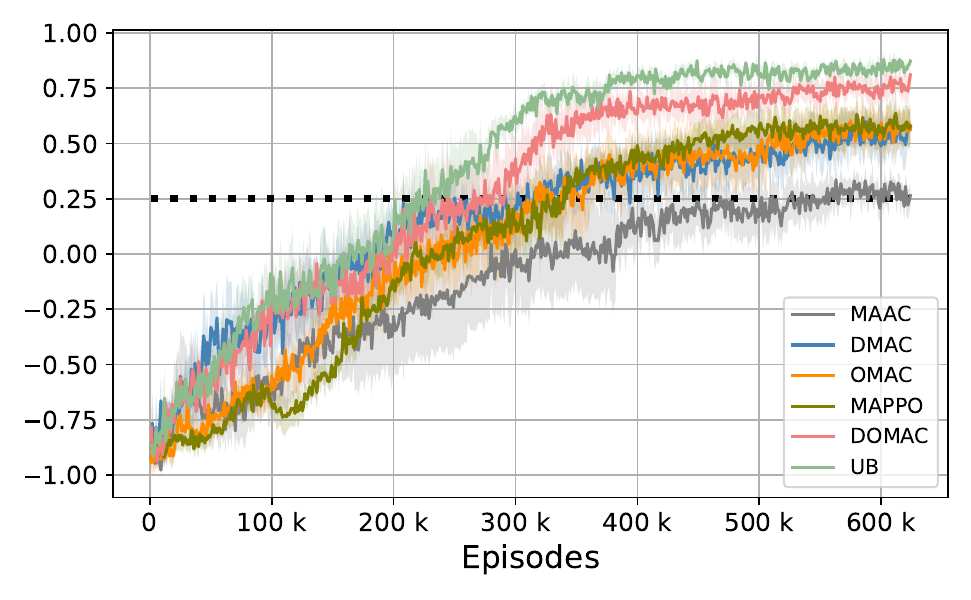}\label{5:d}} 
   \vspace{-0.5em}
\caption{Performance of DOMAC and baselines in the Predator-prey and Pommerman environments.}
\label{fig5}
\vspace{-0.5em}
\end{figure}

\noindent \textbf{Algorithm configuration.}

\noindent For \texttt{PP-3v1} and \texttt{PP-5v2}, the $\rho_{\theta_i, \psi_i}$ and $G_{\phi_i}$ of each controlled agent $i$ are both multi-layer perceptrons (MLP) with $3$ hidden layers of dimensionality $64$. We train the networks for $T=35,600$ episodes using a single environment. The parameters of the networks are updated by $\texttt{Adam}$ optimizer \cite{kingma2015adam} with learning rate for $\rho_{\theta_i, \psi_i}$ and $G_{\phi_i}$ as
$2.5e-4$ and $1e-4$. Rather than updating the networks for every $T_f$ steps, 
we update the networks with data collected from entire $10$ episodes because the predator-prey environment consumes less memory for storing data. We set the number of quantiles as $K=5$, the discounting factor as $\gamma = 0.95$, and the entropy coefficient as $\alpha_i = 0.01$. We have conducted a study on the impact of sample size $l$. We observe that when $l$ is small, increasing it improves the performance obviously. However, a large sample size only brings marginal benefits while requiring too much computation. {Therefore, we set the sample size of \texttt{PP-5v2} as $l=10$.}  %
The others follow the default setting of Pytorch \cite{paszke2017automatic}. 

The configuration for \texttt{Pomm-FFA} and \texttt{Pomm-Team} is generally the same as that of the predator-prey games. However, here $\rho_{\theta_i, \psi_i}$ and $G_{\phi_i}$ are both convolutional neural networks (CNN) with $4$ hidden layers, each of which has $64$ filters of size $3 \times 3$, as the observations are image-based. Between any two consecutive CNN layers, there is a two-layer MLP of dimension $128$. {The learning rate for $\rho_{\theta_i, \psi_i}$ and $G_{\phi_i}$ are both $2.5e-5$.} We parallel $16$ environments during training and the number of forward step is $T_f = 5$, that is, we update the networks after collecting $5$ steps of data from $16$ environments at each iteration. The total number of training episodes is $T = 624,000$. {We set the sample size $l$ in \texttt{Pomm-FFA} and \texttt{Pomm-Team} as 80 and 25 separately.}


{For environments in SMAC, $\rho_{\theta_i, \psi_i}$ and $G_{\phi_i}$ are both MLP of three hidden layers with dimension 128. The discount factor $\gamma$ is set to 0.99. The target networks are updated at a frequency of every 100 episodes. The learning rate for $\rho_{\theta_i, \psi_i}$ and $G_{\phi_i}$ are both $5e-4$. The training steps of \texttt{5m\_vs\_6m} and \texttt{2s3z} are $T=150,0000$, while for \texttt{mmm2} and \texttt{1c3s5z} it extends to $T=2,000,000$. Periodic evaluations are conducted every $10,000$ step to assess the models' performance. 
Episodes are finished upon the defeat of an entire army or upon reaching a designated time limit. We set the sample size as $l=323$ for \texttt{5m\_vs\_6m} and \texttt{2s3z} and set $l=500$ in \texttt{mmm2} and \texttt{1c3s5z}.} All experiments are carried out in a machine with Intel Core i9-10940X CPU and a single Nvidia GeForce 2080Ti GPU. 

\subsection{Experiment Results}
\subsubsection{Comparison with Baseline Methods}
{The performance of DOMAC and baseline methods is evaluated based on two criteria: the average return, indicating effectiveness, and the learning speed, reflecting the rate at which proficiency is attained by each method.}

\noindent \textbf{Analysis of the average return.}

\begin{figure}[htp]
\vspace{-1.0em}
   \centering
    \subfigure[\texttt{5m\_vs\_6m}.]{\includegraphics[width=.22\textwidth]{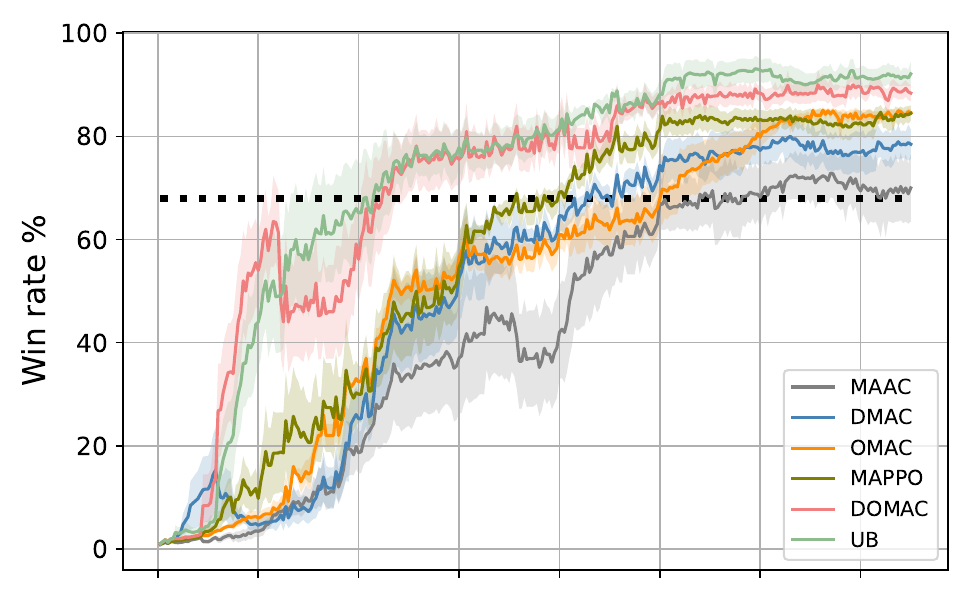}\label{6:a}} \vspace{-1mm}
     \subfigure[\texttt{mmm2}.]{\includegraphics[width=.22\textwidth]{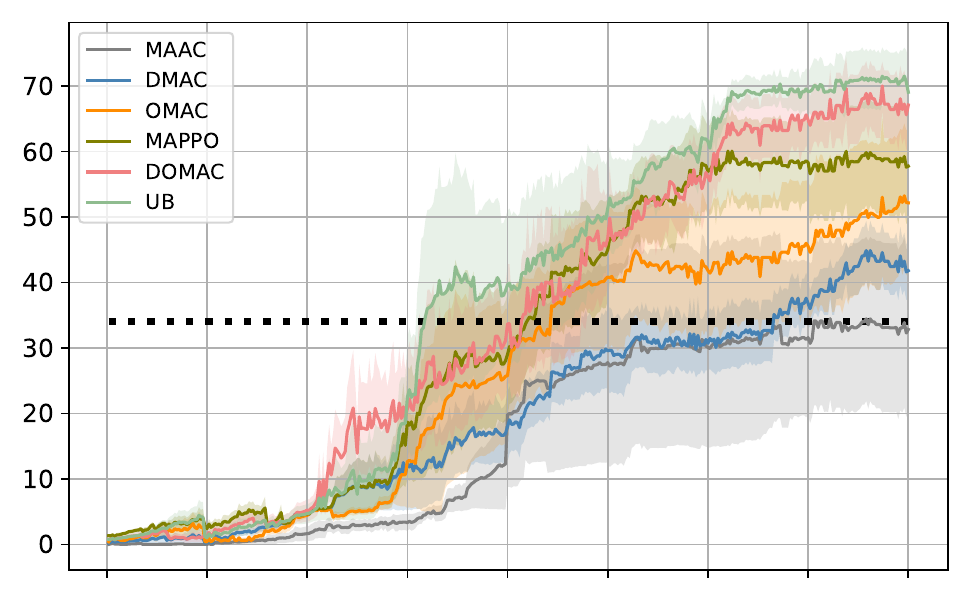}\label{6:b}} \vspace{-1mm} \\
     \subfigure[\texttt{2s3z}.]{\includegraphics[width=.22\textwidth]{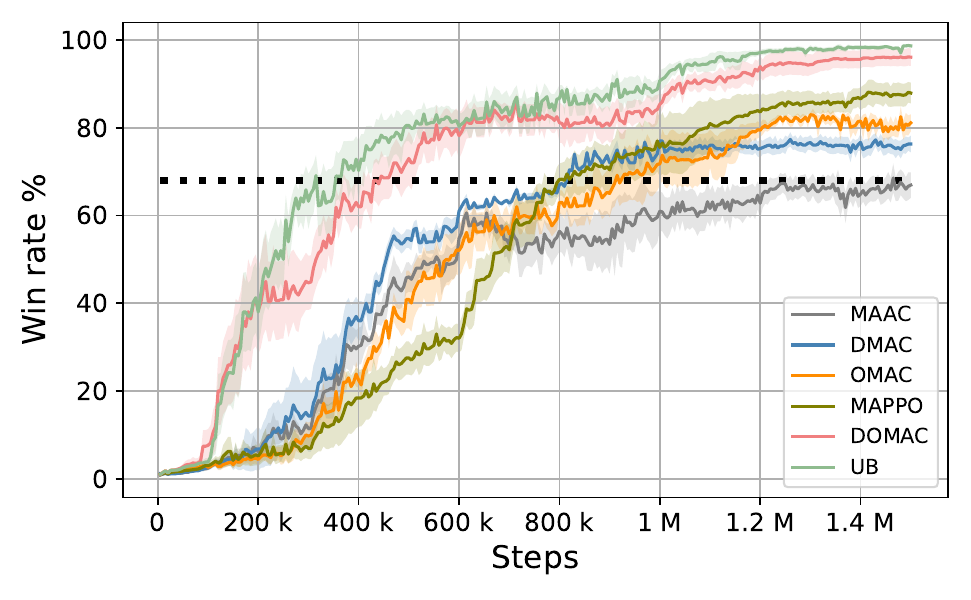}\label{6:c}} 
     \subfigure[\texttt{1c3s5z}.]{\includegraphics[width=.22\textwidth]{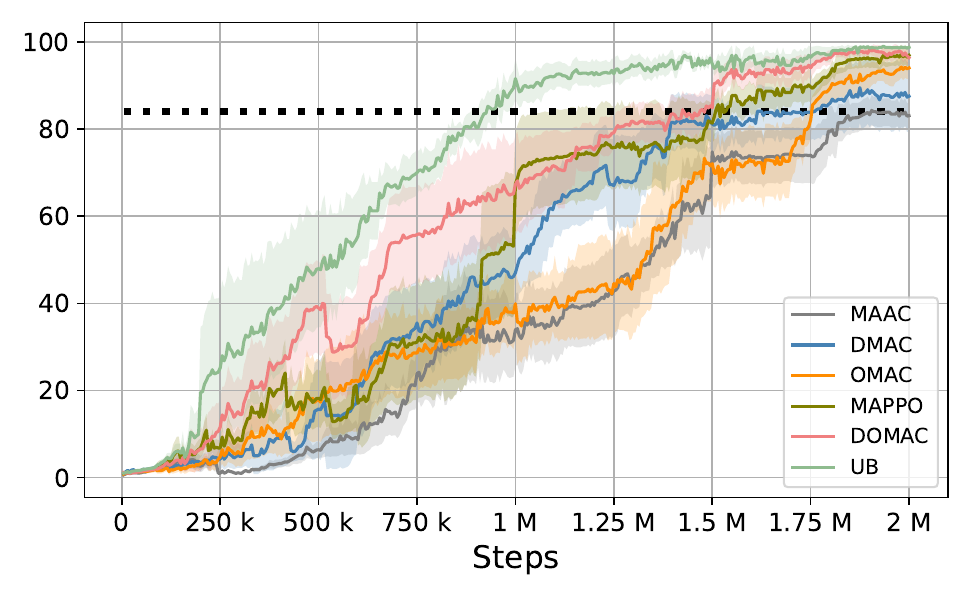}\label{6:d}} \vspace{-1mm}
   \vspace{-0.5em}
\caption{Performance of DOMAC and baseline in the four evaluation scenarios of SMAC.}
\label{fig6}
\vspace{-0.5em}
\end{figure}

\noindent We first compare the overall performance of DOMAC with baselines.
The results are summarized in {Figure \ref{fig5} and Figure \ref{fig6}.}  {{For the} Predator-prey and Pommerman {environments},} we {report the} average returns {for} 8 random seeds. {For SMAC, we report the average win rate.} Specifically, for \texttt{PP-3v1} and \texttt{PP-5v2}, {we evaluate all the methods with $100$ test episodes after every $100$ iterations of training,} and report the mean (solid lines) and the standard deviation (shaded areas) of the average returns over eight seeds. Similarly, for \texttt{Pomm-FFA} and \texttt{Pomm-Team}, all methods are evaluated with $200$ test episodes after every $1,000$ training iteration. {For the scenarios of SMAC, the methods are evaluated with $32$ test episodes after every $5,000$ training steps. We report the mean test win rate (percentage of episodes won {by controlled team}) along with one standard deviation of the test win rate (shaded {areas} in figures).}

{From the curves, it is evident that DOMAC's performance closely aligns with the upper bound while it consistently surpasses the other four benchmarks, achieving faster convergence and demonstrating lower variability in all tested games. This reinforces the effectiveness of DOMAC in forming dependable speculative opponent models without relying on access to actual opponent information. The success of OMAC highlights the benefits of integrating opponent models into the actor-critic framework augmented with a conventional centralized critic, akin to MAAC. Additionally, the superior performance of DMAC over MAAC underscores the value of converting a standard critic into a distributional one, which in turn facilitates the development of more robust policies. These findings individually point to the merits of both Opponent Modelling Advantage (OMA) and Centralized Distributional Critic (CDC) in enhancing performance. Furthermore, when compared to MAPPO~\cite{yu2022surprising}, a state-of-the-art policy gradient algorithm, DOMAC outshines in both sample efficiency and overall performance. This suggests that learning distinct return distributions and accurately anticipating the actions of unknown opponents can indeed significantly elevate performance.}

\noindent {\textbf{Analysis of the learning speed.}} 

\noindent {To demonstrate the efficiency of our approach, we compare the relative learning speed of our methods and baselines with that of MAAC (without loss of generality). This evaluation is defined by the formula $LS = {E_{P_{MA}} }/T$, where $P_{MA}$ represents the optimal performance for MAAC (indicated by the black dashed line in Figures \ref{fig5} and \ref{fig6}), and $E_{P_{MA}}$ denotes the episode count at which different methods reach this benchmark performance (including MAAC, DMAC, OMAC, MAPPO, DOAMC, and UB). To give an example, in the scenario \texttt{PP5v2}, with a total of $T=356000$ training episodes, DOMAC achieves the same performance as $P_{MA}$ at episode 107500, resulting in a relative learning speed of $30.2\%$. As summarized in Table~\ref{T1}, the learning speed of DOMAC consistently surpasses all other methods by a large margin over all the tested scenarios. Furthermore, it is worth noting that DOMAC exhibits a convergence speed comparable to UB (the baseline trained with ground-truth opponents' information). The exceptional data efficiency of DOMAC can be attributed to the distributional critic's role in effectively guiding the opponent modelling process. Meanwhile, the opponent model, in turn, aids the agent's policy in making more informed decisions. In contrast, DMAC and OMAC are inferior due to the lack of guidance from the opponent models and the distributional critic, respectively.}


\begin{table}[!h]
\setlength\tabcolsep{0.6pt}
\caption{{The learning speed of different methods.}}
\vspace{-0.8em}
\begin{tabular}{lllllll}
\hline
          & MAAC & DMAC & OMAC & MAPPO & DOMAC&UB \\ \hline
\texttt{PP3v1}     &   77.4\%   &  36.7\%    &   45.2\%   &    35.3\%   & \textbf{31.1\%} &  \textbf{29.9}\%    \\ \hline
\texttt{PP5v2}     &   77.1\%   &   42.4\%   &  45.2\%    &   39.5\%    & \textbf{30.2\%}  &  \textbf{16.9\%}  \\ \hline
\texttt{FFA}  &  76.4\%    &   48.1\%   &   52.4\%   &   41.8\%    &  \textbf{39.1\%}   &   \textbf{35.8\%}    \\ \hline
\texttt{Team} &    59.1\%  &  38.5\%    &  42.6\%    &  37.9\%     &  \textbf{34.5\%}  &  \textbf{32.1\%}\\ \hline
\texttt{5m\_vs\_6m}        &  72.0\%    &   56.3\%   &  66.3 \%   &  47.3\%   &  \textbf{29.3\%}  & \textbf{27.6\%}  \\ \hline
\texttt{2s3z}      &   82.7\%   &  54.1\%    &   53.5\%   &  35.3\%     & \textbf{31.6\%}  &  \textbf{28.4\%}\\ \hline
\texttt{mmm2}      &   88\%   &  82\%    &   55\%   &  53\%     & \textbf{51.5\%}  &  \textbf{39.4\%}\\ \hline
\texttt{1c3s5z}      &   90\%   &  80.5\%    &   87.5\%   &  75.2\%     & \textbf{70.1\%}  &  \textbf{46.7\%}\\ \hline

\end{tabular}\label{T1}
\end{table}

\begin{figure}[htp]
\vspace{-5pt}
    \centering
    \includegraphics[width=0.4\textwidth]{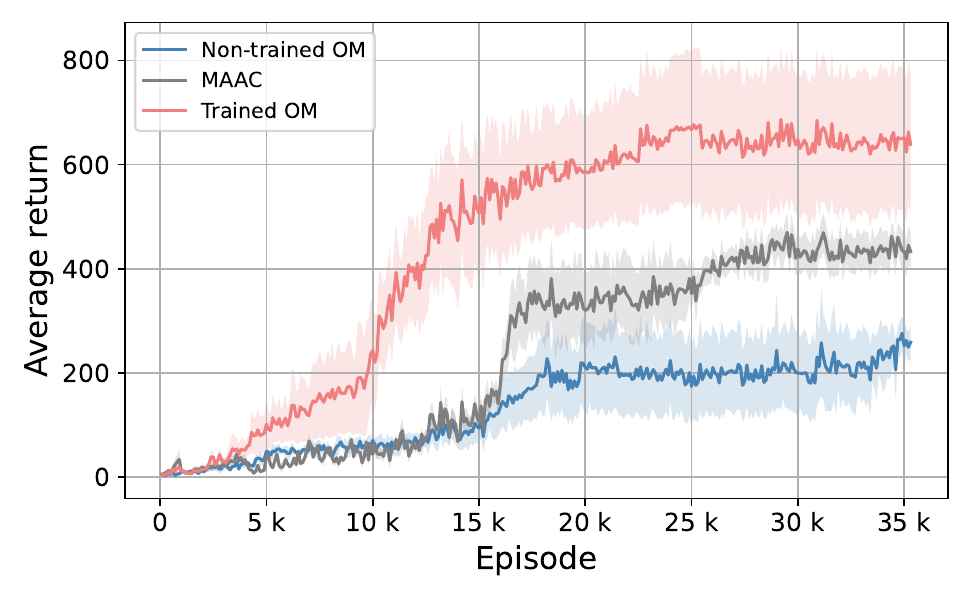}
    \vspace{-5pt}
    \caption{Impacts of the OMA for game  \texttt{PP-5v2}.}
    \label{fig7}
\vspace{-3pt}
\end{figure}

    

\subsubsection{{Ablation studies}}

\noindent{The ablation studies serve to answer the following questions: (a) Does the OMA and CDC help to learn a better policy? (b) How do the hyperparameters in DOMAC affect the final results?}


\noindent \textbf{Ablation studies for OMA.} We first verify that the speculative opponent models truly help to learn a better policy. We perform a pair of experiments: the first one uses trained and fixed opponent models and trains the actor from scratch while the second one uses randomly initialized and fixed opponent models instead. The average returns are plotted in {Figure \ref{fig7}.} It is obvious that the DOMAC with trained opponent models learns faster than the one without, which indicates that the agent can infer the {behaviours} of its opponents and take advantage of this knowledge to make better decisions, especially at the beginning stage of the learning procedure. 
It shows that our opponent models can provide reliable information for better decision-making.

\begin{figure}[!t]
\footnotesize
    \centering
    \subfigure[]{\includegraphics[width=.22\textwidth]{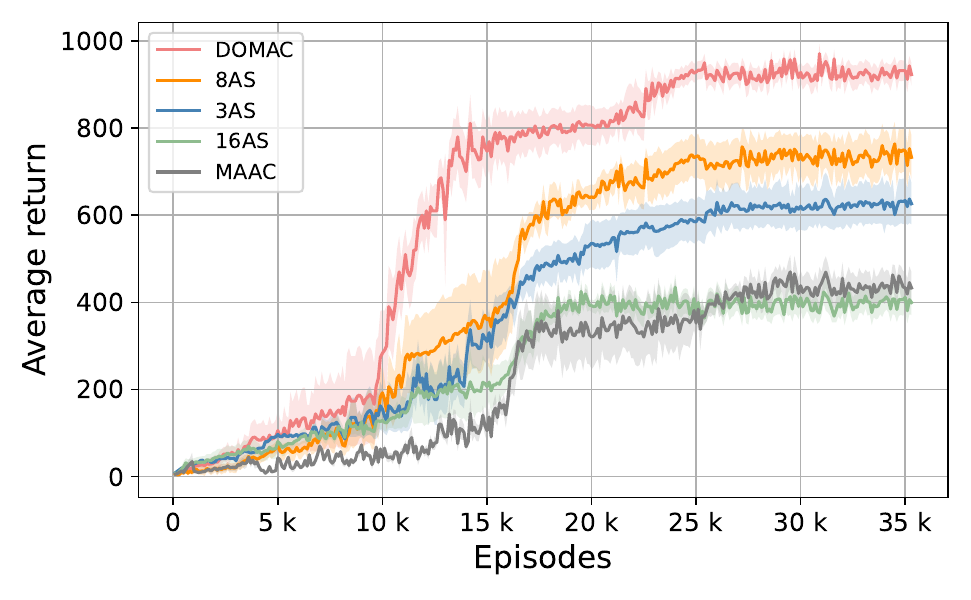}\label{8a_AS}}
    \subfigure[]{{\includegraphics[width=.22\textwidth]{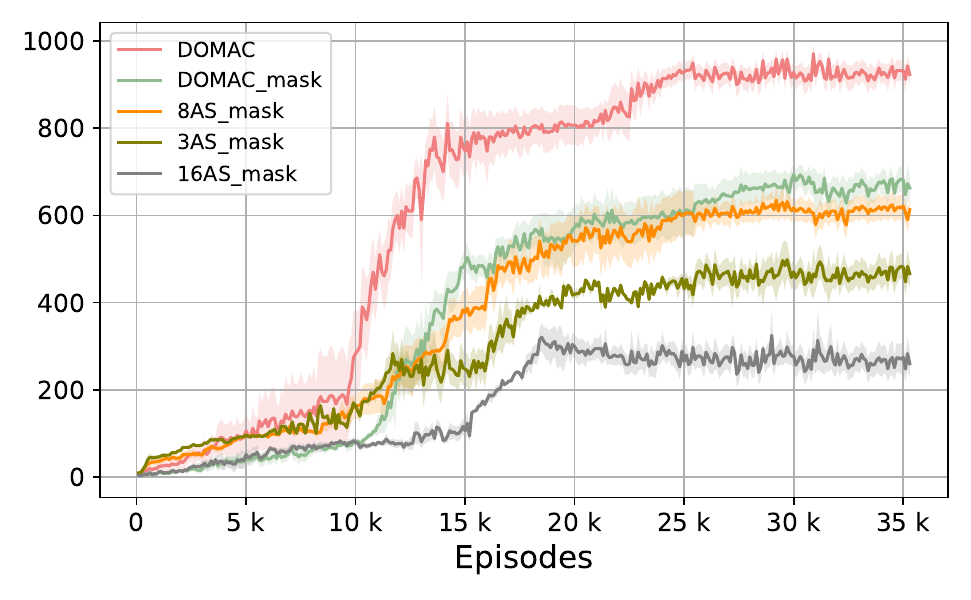}}\label{8b_mask}}
  
    \caption{(a) Study of OMA action space in \texttt{PP-5v2} ,  (b) Study of masked observation in \texttt{PP-5v2}.}
    \label{fig8}
\end{figure}

While the previous results show that our opponent models improve the decision-making quality, one may wonder whether the improvement really results from the opponent action prediction output by the opponent models. Is it possible that the actor can always get the improvement by conditioning on arbitrary information output by the opponent models? To investigate this question, we change the output dimension $d$ of opponent models to 3, 8, and 16 respectively while retaining the other configurations. We denote these setting as ``3AS'', ``8AS'', and ``16AS'' in {Figure \ref{8a_AS}.} Note that the opponent action space size is 5. Therefore, the opponent models in the three new settings output some conditional information instead of opponent action predictions.
The training results over 8 random seeds are shown in {Figure \ref{8a_AS}}.
We can see that the original DOMAC learns faster and better than the other ones. We also notice that the performance of $d=3$ and $d=8$ are consistently better than that of $d=16$ for the entire training period. It implies that the opponent model can infer more reliable information if its output dimension is close to the opponent action space size.  

Given the above results, the following question is that why the opponent models perform the best when their architectures are designed to output opponent action prediction. A possible reason is that the opponents sometimes appear in the observations of the controlled agents and thus, the local observations can occasionally convey the state-changing information of the opponents.
In this case, the opponent models that output opponent action prediction can indeed learn better from the received observations. To verify this hypothesis, we conduct another experiment where we mask out the opponent information when the agent observes the opponents and retains the other configurations. The results are denoted as ``X\_mask'' where ``X'' means the original algorithm setting. 
{Figure \ref{8b_mask}} shows that when masking out the opponent information, the performance of all algorithms declines,``DOMAC\_mask'' is weak to ``DOMAC'', 
which means that the opponent model learning indeed benefits from the information contained in the local observations. 

\begin{figure}[htp]
\vspace{-1.0em}
   \centering
    \subfigure[ \texttt{PP-5v2}. ]{\includegraphics[width=.22\textwidth]{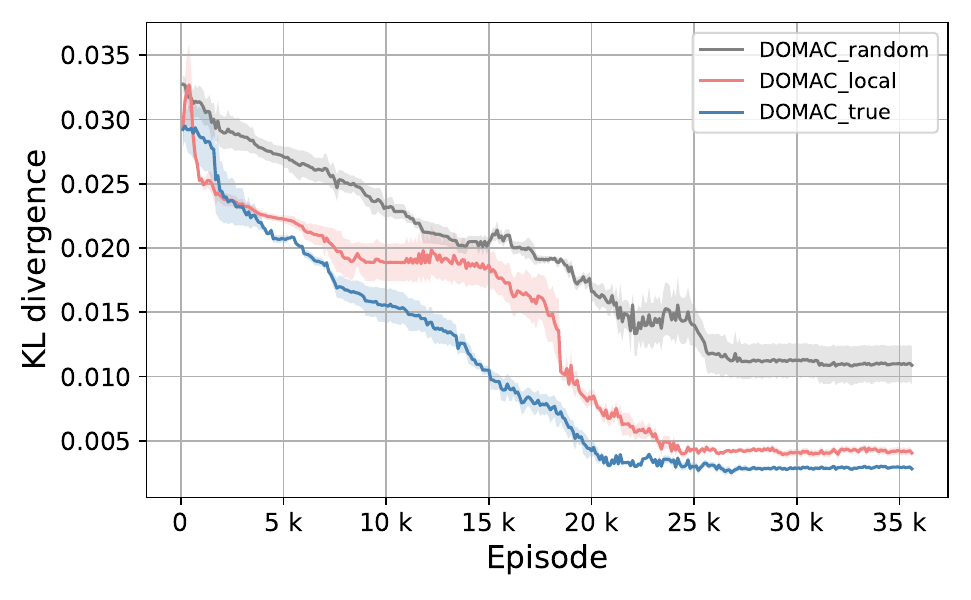}\label{9:a}} \vspace{-1mm}
    \subfigure[\texttt{Pomm-FFA}.]{\includegraphics[width=.22\textwidth]{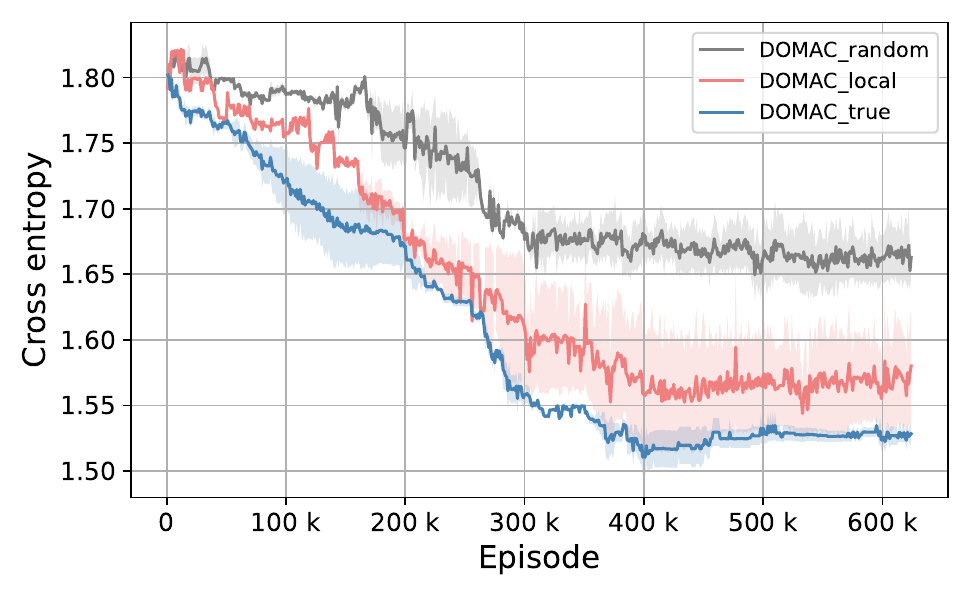}\label{9:b}} \vspace{-1mm}
    
\caption{Study of the influence of the local information.}
\label{fig9}
\vspace{-0.8em}
\end{figure}

{To further illustrate that the opponent models can learn reliable opponent {behaviours} using the local information, we compare our method with the upper bound that directly uses the {ground-truth} opponent information to infer the opponent's {behaviours}. We use game \texttt{PP-5v2} and \texttt{Pomm-FFA} as {the test beds. The} results over eight random seeds are shown in Figure \ref{fig9}. Specifically, we perform a pair of experiments, where the first one uses the real opponent observation to infer the opponent's {behaviours}, while the second one uses randomly generated information to estimate the opponent's {behaviours}.  We measure the performance of different settings by computing the distance with the actual opponent's {behaviours}.  In Figure \ref{fig9}, it is obvious that the DOMAC with local information has a similar performance to the one with true information and {significantly outperforms} the one with randomly generated information. These results show that our opponent models can effectively reason about the opponents' intentions and {behaviours} even with local information.}

\begin{figure}[htp]
\vspace{-5pt}
    \centering
\includegraphics[width=0.4\textwidth]{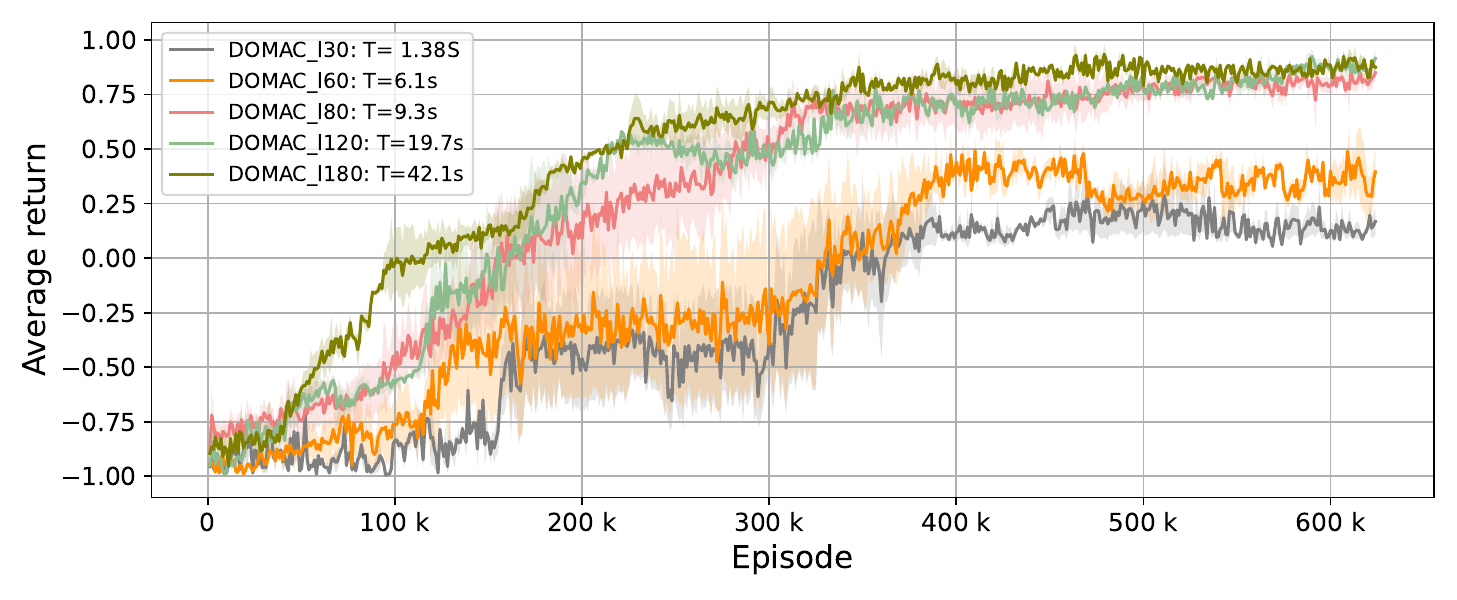}
    \vspace{-5pt}
    \caption{The study of the sample size.}
    \label{fig10}
\vspace{-1pt}
\end{figure}

{{Here we conduct} another experiment to {examine} the impact of {sampling} size {$l$}. We take \texttt{Pomm-FFA} as a demonstration. Given that there are three opponents in \texttt{FFA} and the size of the action space is $6$, the size of the joint opponent action space is $216$. We consider the {sampling size $l$ to be $60$, $80$, $120$, and $180$, respectively and keep the rest of the configurations fixed.} In Figure \ref{fig10}, {we observe that the performance gradually improves with the sampling size. The performance of $l=80$ is comparable to that of $l= 120, 180$}. We also {report} the {inference} time of DOMAC under different {sampling} sizes. {Specifically,} $T_{30}=1.38s, T_{60}=6.1s$, $T_{80}=9.3s$, $T_{120}=19.7s$ and $T_{180}=42.1s$. Therefore, we can conclude that a sample size larger than 100 only brings marginal benefits while {costing} too much computation. To balance efficiency and performance, we choose $l=80$ in our experiment.} 

\begin{table*}[!h]\tiny
\caption{The average evaluation results and running time for different methods on various quantile numbers}
\vspace{-2.0em}
\begin{center}
\begin{tabular}{l|lc|ll|cc|ll}
\hline
Env      & \multicolumn{2}{l|}{K=3}            & \multicolumn{2}{l|}{K=5}                                 & \multicolumn{2}{c|}{K=10}           & \multicolumn{2}{l}{K=15}                                \\ \hline
         & \multicolumn{1}{l|}{Return} & Time & \multicolumn{1}{l|}{Return} & \multicolumn{1}{c|}{Time} & \multicolumn{1}{l|}{Return} & Time & \multicolumn{1}{l|}{Return} & \multicolumn{1}{c}{Time} \\ \hline
\texttt{PP3v1}    & \multicolumn{1}{l|}{$651.1 \pm 13.6$}        &  $4.7 \pm 1.3$ (s)   & \multicolumn{1}{l|}{$708.1\pm 28.9$}      &     $8.4\pm 1.1$(s)                      & \multicolumn{1}{c|}{$816.1\pm 16.1$}        &   $21.2 \pm 2.5$ (s)   & \multicolumn{1}{l|}{$754.8 \pm 36.1$}        &    $41.15 \pm 8.1$(s)                      \\ 
\texttt{PP5v2}    & \multicolumn{1}{l|}{$857.3 \pm 15.2$}        &  $6.4 \pm 0.9$(s)    & \multicolumn{1}{l|}{$938.4\pm 32.9$}        &     $13.7\pm 1.2$(s)                      & \multicolumn{1}{c|}{$969.1\pm 26.1$}        &   $23.4 \pm 3.0$ (s)   & \multicolumn{1}{l|}{$974.8 \pm 31.1$}        &    $44.31 \pm 2.7$(s)                      \\ \hline
\texttt{FFA} & \multicolumn{1}{l|}{$0.783 \pm 0.6$}        &    $7.2 \pm 2.1$(s)  & \multicolumn{1}{l|}{$0.842 \pm 0.37$}        &    $10.5 \pm 1.5$(s)                      & \multicolumn{1}{c|}{$0.871 \pm 0.3$}        &  $26.1 \pm 2.3$(s)    & \multicolumn{1}{l|}{$0.91 \pm 0.3$}        &       $42.8 \pm 8.1$(s)                  \\ 
\texttt{Team} & \multicolumn{1}{l|}{$0.817 \pm 0.35$}        &    $10.1 \pm 2.4$(s)  & \multicolumn{1}{l|}{$0.871 \pm 0.41$}        &    $18.1 \pm 1.9$(s)                      & \multicolumn{1}{c|}{$0.886 \pm 0.12$}        &  $35.8 \pm 3.7$(s)    & \multicolumn{1}{l|}{$0.94 \pm 0.23$}        &       $55.1 \pm 6.3$(s)                  \\ \hline
\texttt{5m\_vs\_6m}     & \multicolumn{1}{l|}{$98.3  \pm 0.1\%$}        & $7.8 \pm 1.4$(s)     & \multicolumn{1}{l|}{$99.1 \pm 0.01\%$}        &     $15.2 \pm 2.5$(s)                      & \multicolumn{1}{c|}{$99.3\pm 0.01\%$}        &    $30.77 \pm 4.3$(s)  & \multicolumn{1}{l|}{$99.8 \pm 0.01\%$}        &       $48.01 \pm 8.8$(s)                   \\
\texttt{2s3z}     & \multicolumn{1}{l|}{$96.1  \pm 0.5\%$}        & $8.1 \pm 1.3$(s)     & \multicolumn{1}{l|}{$98.2 \pm 0.06\%$}        &     $14.8 \pm 1.1$(s)                      & \multicolumn{1}{c|}{$98.9\pm 0.8\%$}        &    $31.25 \pm 2.1$(s)  & \multicolumn{1}{l|}{$99.1 \pm 0.01\%$}        &       $40.28 \pm 10.2$(s)                   \\ \hline
\texttt{mmm2}     & \multicolumn{1}{l|}{$53.1  \pm 1.6\%$}        & $13.7 \pm 2.1$(s)     & \multicolumn{1}{l|}{$60.1 \pm 3.2\%$}        &     $20.7 \pm 4.7$(s)                      & \multicolumn{1}{c|}{$63.4\pm 2.5\%$}        &    $35.27 \pm 3.8$(s)  & \multicolumn{1}{l|}{$68.8 \pm 4.3\%$}        &       $44.01 \pm 7.3$(s)                   \\
\texttt{1c3s5z}     & \multicolumn{1}{l|}{$93.1  \pm 1.1\%$}        & $7.2 \pm 2.5$(s)     & \multicolumn{1}{l|}{$96.9 \pm 0.1\%$}        &     $19.8 \pm 1.1$(s)                      & \multicolumn{1}{c|}{$97.1\pm 0.5\%$}        &    $30.13 \pm 3.2$(s)  & \multicolumn{1}{l|}{$97.3 \pm 0.1\%$}        &       $45.29 \pm 12.1$(s)                   \\ \hline
\end{tabular}
\end{center}\label{T2}

\end{table*}

\noindent{\textbf{Ablation studies for CDC.} 

\noindent{{In} CDC, we use the quantile representation technique to approximate the return distribution. The number of quantiles may influence the algorithm's performance. We conduct an ablation study for different values of $K$ under different scenarios to further reveal the influence of the quantile number $K$ {during training}. We {variate} the quantile {number} $K$ to $3$, $5$, $10$, and $15$ respectively while {fixing} other configurations. Here, we {present the training curves of our method for} games \texttt{PP-5v2} and \texttt{2s3z} {(Figure~\ref{fig11})}. We can see that the performance of $K=5,10,15$ quantiles is consistently better than that of $K=3$ quantiles. This is aligned with the intuition that more quantiles can support more fine-grained distribution {modelling}. Thus, it better captures the return randomness and improves performance. {The results for the rest {scenarios} are in the appendix (Figure A1).}} {{For testing, we summarize the results of the total eight scenarios in Table \ref{T2}}. We show the average return and running time over eight seeds. In the Predator-prey and Pommerman environments, the data was gathered by averaging the returns and calculating the evaluation times at the checkpoint, which achieved the highest average performance during training. In the scenarios of SMAC, we report the median of the final ten evaluation win rates and evaluation times\cite{wang2021rode}. The results show that the performance improved with the increase of the quantile number. Nevertheless, it is noteworthy that the computational time exhibits an increasing trend as the number of quantiles expands.  To balance computational overhead and performance, we set $K=5$ throughout the experiments.} 
\begin{figure}[htp]
\vspace{-1.0em}
   \centering
    \subfigure[ \texttt{PP-5v2}. ]{\includegraphics[width=.22\textwidth]{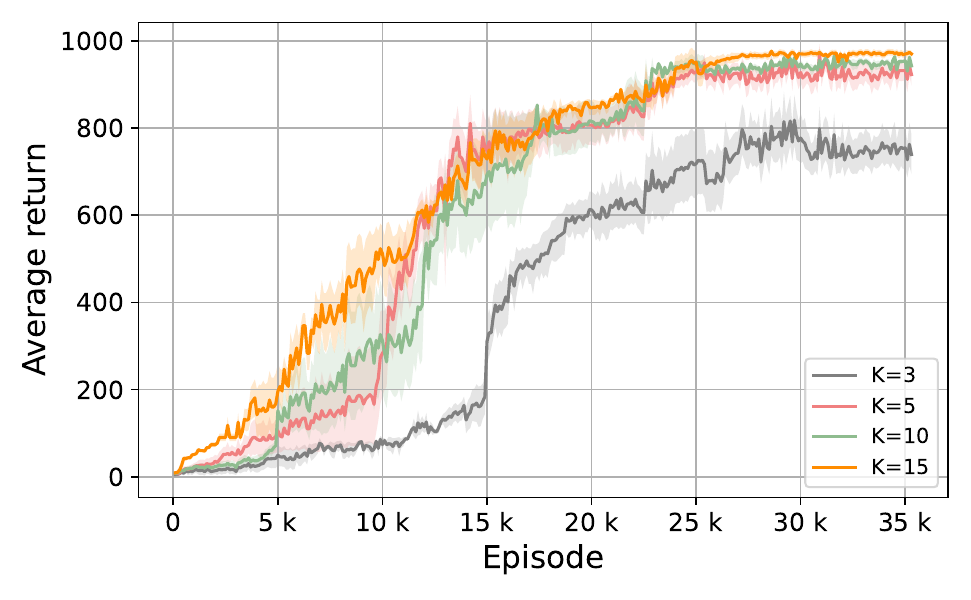}\label{11:a}} \vspace{-1mm}
    \subfigure[\texttt{2s3z}.]{\includegraphics[width=.22\textwidth]{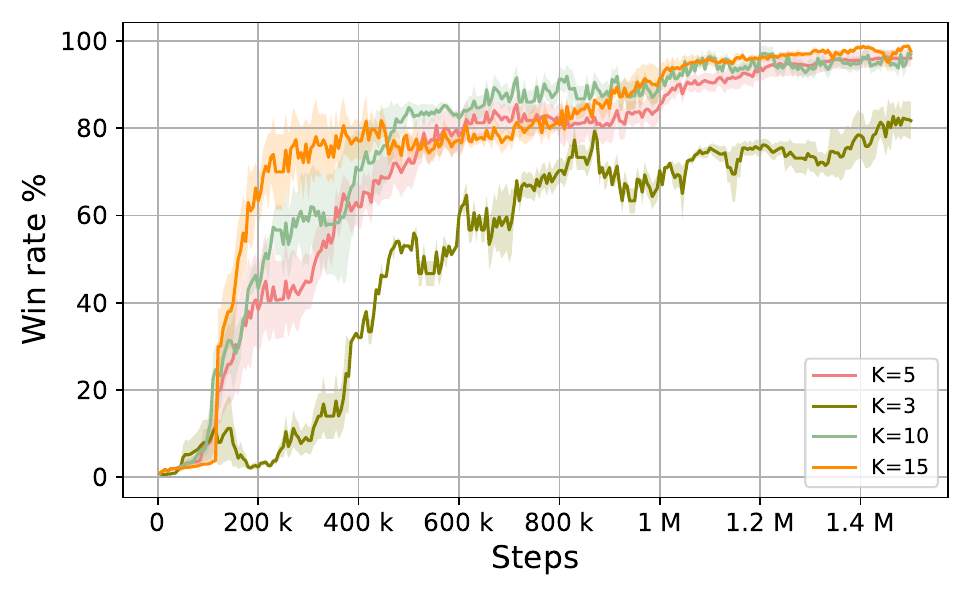}\label{11:b}} \vspace{-1mm}
    
\caption{{Study of the impact of the quantile numbers.}}
\label{fig11}
\end{figure}

\subsubsection{Exposing connections between OMA and CDC}
We are curious about how the centralized distributional critic (CDC) impacts the learning of the opponent models and policy. {To answer this question, we show that, with the help of the CDC, the opponent models are more confident in predicting opponents' actions, and the predictions are also more accurate. {In turn}, the CDC can assist the policy in identifying actions that yield more rewards, which results in better policies. Here, we also use the game \texttt{PP-5v2} for demonstration. The results for the rest games are in the appendix (Figure \ref{figa1}). In the following, we provide evidence for each argument, respectively.} 

\noindent \textbf{{Opponent models are more accurate with CDC.}} 

\noindent To prove that the opponent models are more accurate in predicting opponent actions with CDC, we adopt two metrics: 1). We compute the average entropy of the predicted probability distribution over opponent actions, which measures the confidence of the prediction. The less average entropy is, the more certain are the agents about the predicted opponent actions.  2). We compute the Kullback-Leibler (KL) divergence \cite{kullback1951information} between the predicted and the true probability distribution over opponent actions. The {KL divergence} is the direct measure of the distance between the opponent models and the true opponent policies. The less the {KL divergence} is, the more similar are the opponent models to the true policies. 
Note that we only use the true opponent policies for evaluation. We do not use them to train DOMAC.
From Figures \ref{12:a} and \ref{12:b}, we conclude that the CDC can increase the training speed (faster descent) and improve the reliability
and confidence (lower {KL divergence} and entropy) of the opponent models. Note that for the pommerman environment, we can not access the opponents' policies but their actions instead due to the environment restrictions. Therefore, in \texttt{Pomm-FFA} and \texttt{Pomm-Team} games, we replace {KL divergence} with cross entropy loss \cite{zhang2018generalized} which has a similar effect. The second argument, i.e., the CDC helps policy to identify actions with more rewards, is supported by the results in {Figure \ref{fig5} and \ref{fig6}}, where the performance of DMAC is better than MAAC. Note that the expected returns of DMAC are still lower than DOMAC, which implies that the integration of OMA and CDC is essential for our method.

\noindent\textbf{CDC helps policy to identify actions with more rewards.}

\noindent {This argument is supported by the results that DOMAC has the lowest entropy in Figure \ref{12:c} and highest average episode rewards in Figure \ref{fig5} and Figure \ref{fig6}. Furthermore, note that both OMAC and DMAC can also obtain a policy with a very low entropy after training. However, their expected returns are still lower than DOMAC, which means the actions they output with high confidence are not as good as those output by DOMAC. This implies that the integration of OMA and CDC is essential for our algorithm.}

\begin{figure*}[htp]
\vspace{-1.5em}
    \centering
    
    \subfigure[Average entropy of OM.]{\includegraphics[width=.3\textwidth]{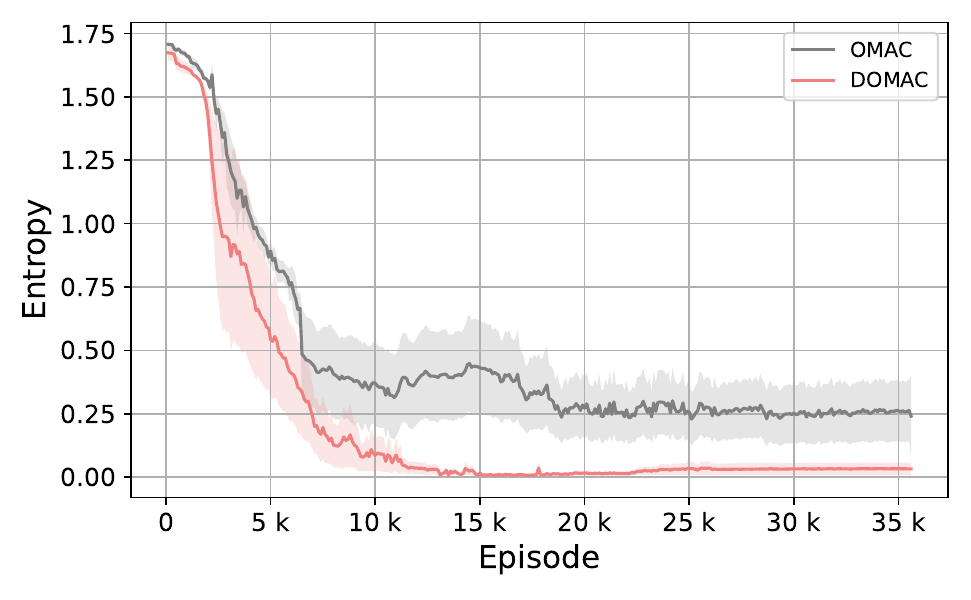}\label{12:a}}
    \subfigure[Average KLD of OM.]{\includegraphics[width=.3\textwidth]{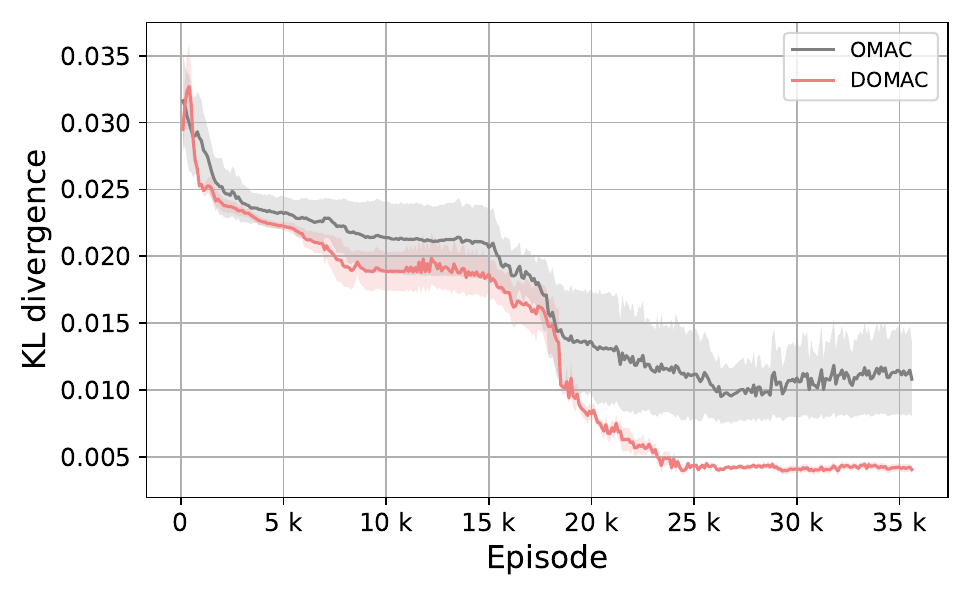}\label{12:b}}
\subfigure[Average entropy of policies.]{\includegraphics[width=.3\textwidth]{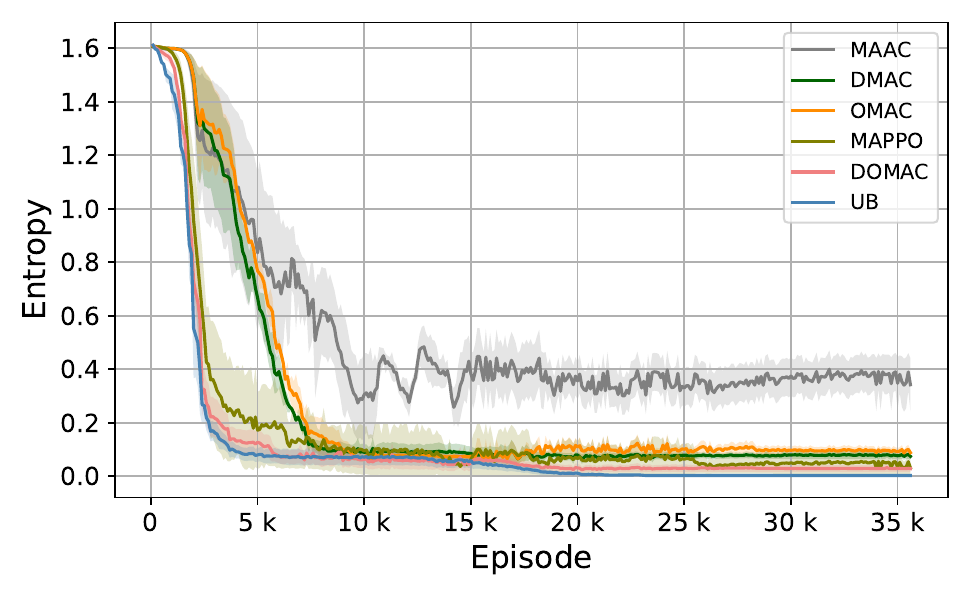} \label{12:c}}
\vspace{-0.5em}
\caption{Impacts of the distributional critic on opponent models (OM) and policies for game  \texttt{PP-5v2}.}
\label{fig12}
\vspace{-1.0em}
\end{figure*}

\section{Conclusion and Future Work}\label{s6}
This paper proposes a distributional opponent model aided actor-critic (DOMAC) algorithm by incorporating the distributional RL and speculative opponent models into the actor-critic framework. In DOMAC, the speculative opponent models take as input the controlled agents' local observations, which realizes the opponent {modelling} when opponents' information is unavailable. 
With the guide of the distributional critic, we manage to train the actor and opponent models effectively.
Extensive experiments demonstrate that DOMAC not only obtains a higher average return but also achieves a faster convergence speed. The ablation studies and the baselines OMAC and DMAC prove that the OMA and CDC are both essential parts for our algorithm. That is, the CDC leads to a higher-quality OMA and in turn, the better OMA helps to improve the overall performance. Based on our above analysis, the proposed integration of OMA and CDC has the potential to improve the performance of various RL algorithms. 
{For future work, we will study how to model
opponents with dynamic strategies only using the local information.  In addition, we plan to {investigate extending DOMAC to value-based methods}. By doing so, we will show that the integration of distributional reinforcement learning and opponent {modelling} can act as a general plugin that benefits a wide range of reinforcement learning algorithms.}

\medskip

\bibliography{aaai23}

\clearpage

\appendix

\providecommand{\upGamma}{\Gamma}
\providecommand{\uppi}{\pi}
\section{Distributional opponent model aided policy gradient theorem} \label{proof_thm_1}

To clarify the idea of this work, we first compare the traditional MARL algorithms with the existing works on opponent \colorb{modelling}. Specifically, the traditional MARL algorithms use the information (observations and actions) of the controlled agents to train the team policy. These algorithms normally consider the opponents as a part of the environment. In comparison, the existing works on opponent \colorb{modelling} use opponent models to predict the goal/actions of the opponents so that the controlled team can make better decisions given those predictions. However, these opponent \colorb{modelling} works need to use the true information of the opponents to train the opponent models. 
In this work, we don't use the opponents' observations and actions to train the opponent model directly because we consider the setting where the opponent's information is inaccessible. That means our work tries to maintain the advantages brought by the opponent models while using the same information as the traditional MARL algorithms. However, this brings a challenge that we don't have the true opponent actions to act as the training signal of our opponent model. To find an alternative training signal, we propose to use the distributional critic. Because the distributional critic can provide the training signal for the controlled agent's actions while our controlled agent's actions are conditioned on the outputs of opponent models, the distributional critic enable the training of opponent models without true opponent actions.


In this work, we apply the distributional RL to the POMDP setting. Note that most of the existing works focus on the MDP setting, i.e., they use the global states as the input of the centralized critic while we use the joint observations instead. One may have the concern about whether it is feasible to learn with the joint observations. To address this concern,
we would like to first clarify what a state $s$ is. By definition, a state $s$ is one configuration of all agents and the external environment. However, in practice, a state being used by the algorithms often cannot represent the configuration of external environment completely because there are many latent variables that can affect the state transition. For example, in Atari games, a state is an image that represents the current game status while the internal random seed of the simulator can affect the future states and it cannot be obtained by the learning algorithms. Therefore, when a learning algorithm takes a state as input, the state actually means the full information that is available to the algorithm. For the latent variables that the algorithm has no access to, we use the state transition function to depict the randomness brought by them.

In this sense, 
when learning the critic in POMDP setting, it is equivalent to MDP setting as long as the critic takes all available information (i.e., joint observation of all controlled agents) as input. 
Specifically, the return expectation of MDP setting is over the trajectories resulting from the state transition function $P(s'|s,a)$ and agent policies, i.e., $\mathbb{E}_{{s}\sim P(s'|s,a), a\sim \pi} [ \mathbb{E}(Z({s}, {a}))]$. Similarly, we can let the return expectation of POMDP setting to be over the trajectories resulting from the observation transition function $P(o'|o,a)$ and agent policies, i.e., $\mathbb{E}_{{o}\sim P(o'|o,a), a\sim \pi} [ \mathbb{E}(Z({o}, {a}))]$ (Equation (5) in our paper). Based on the state transition function $P(s'|s,a)$ and observation function $\mathcal{O}(o|s)$, we can define the observation transition function as
\renewcommand{\thetable}{13}
\begin{equation}
P(o'|o,a)=\sum_{s}P(s|o)\sum_{s'}P(s'|s,a)\mathcal{O}(o'|s'),
\end{equation}
where $P(s|o)=\frac{P(s,o)}{P(o)}=\frac{P(s)\mathcal{O}(o|s)}{\sum_{s}\mathcal{O}(o|s)P(s)}$ according to the Bayesian theorem. Like in MDP, the observation transition function depicts the randomness brought by all unobservable information. Therefore, the return expectation of MDP and POMDP actually have the same mathematical form and they only use different transition functions to depict the randomness from unavailable information. In both settings, we can sample the trajectories for return estimation without explicitly learning the state transition function or observation transition function. Note that the return estimation in POMDP generally has more uncertainty than the return estimation has in MDP because POMDP has more unavailable information to introduce randomness in the observation transition function. This, however, necessitates the use of distributional critic to better cope with the return estimation uncertainty. 

\section{Environmental settings} \label{Appendix-B}

 \textbf{Predator-prey: }The states, observations, actions, and state transition function of each agent is formulated as below.

 \begin{itemize}
   \item \textbf{The observations.} Observations consist of high-level feature vectors containing relative distances to other agents and landmarks. 
     \item \textbf{Actions space.} Any agent, either predators or preys, has five actions, i.e. $[\texttt{up}, \texttt{down}, \texttt{left}, \texttt{right}, \texttt{no-op}]$ where the first four actions means the agent moves towards the corresponding direction by one step, and $\texttt{no-op}$ indicates doing-nothing. All agents move within the map and can not exceed the boundary. 
     \item \textbf{State transition $\mathcal{T}$.} The new state after the transition is the map with updated positions of all agents due to agents moving in the world. The termination condition for this task is when all preys are dead or for 500 steps.
\end{itemize}

 \textbf{Pommerman:} 

\begin{itemize}
    \item \textbf{The states and observations.} 
    At each time step, agents get local observations within their field of view $5\times 5$, which contains information (board, position,ammo) about the map. The agent obtain information of the Blast Strength, whether the agent can kick or not, the ID of their teammate and enemies, as well as the agent's current blast strength and bomb life.
    \item \textbf{Actions space.} Any agent chooses from one of six actions, i.e. $[ \texttt{up}, \texttt{left}, \texttt{right}, \texttt{down}, \texttt{stop}, \texttt{bomb}]$. Each of the first four actions means moving towards the corresponding directions while \texttt{stop} means that this action is a pass, and \texttt{bomb} means laying a bomb.
    
    \item \textbf{Rewards $\mathcal{R}$.}  In $\texttt{Pomm-Team}$, the game ends when both players on the same team have been destroyed. It ends when at most one agent remains alive in $\texttt{Pomm-FFA}$. The winning team is the one who has remaining members. Ties can happen when the game does not end before the max steps or if the last agents are destroyed on the same turn. Agents in the same team share a reward of 1 if the team wins the game, they are given a reward of -1 if their team loses the game or the game is a tie (no teams win). They only get 0 reward when the game is not finished.
\end{itemize}

{
 \textbf{SMAC Setting:} 
\begin{itemize}
    \item \textbf{The states and observations.}  At each time step, agents receive local observations, which contain information (distance, relative x, relative y, health, shield, and unit type) about the map within a circular area for both allied and enemy units and makes the environment partially observable for each agent. The global state is composed of joint observations, which could be used during training. All features, both in the global state and in individual observations of agents, are normalized by their maximum values.

    \item \textbf{Action Space.} Any agent has the following actions, i.e., $[ \texttt{up}, \texttt{left}, \texttt{right}, \texttt{down}, \texttt{attack}, \texttt{stop}, \texttt{no-op}]$.
     \item \textbf{Rewards.} At each time step, the agents receive a joint reward equal to the total damage dealt to the enemy agents. In addition, agents receive a bonus of 10 after killing each opponent, and 200 after killing all opponents for winning the game. The rewards are normalized to ensure that the maximum cumulative reward achievable in each scenario is around 20.
    
\end{itemize}

}

\section{Evaluation results} \label{Appendix-D}
\subsection{Analysis of the computation complexity}
{Intuitively speaking, training the speculative opponent model with local information seems more complicated and may need more data.} To investigate whether the OMA and CDC can increase the computation complexity in our method, {we provide the computational cost between different algorithms.} We gather the number of floating-point operations (FLOPS) of a single inference and the number of parameters for each method. Table \ref{T3} shows the computation complexity of different algorithms. It shows that DOMAC has the comparable model volume and computational complexity as MAAC.

\renewcommand{\thetable}{A1}
\begin{table}[!ht]
\centering
\caption{Computational complexity in \texttt{Pomm-Team}}
\begin{tabular}{ccc}
\hline
Algorithm & Params(M) & FLOPs(G) \\ \hline
MAAC      & 8.079     & 0.100         \\ 
OMAC      & 8.085     & 0.101         \\ 
DMAC      & 8.094     & 0.1005         \\ 
DOMAC     & 8.100     & 0.1015         \\ \hline
\end{tabular}\label{T3}
\end{table}

\renewcommand{\thefigure}{A2}
\renewcommand{\thefigure}{A1}
\begin{figure*}[!t]
    \centering
    \subfigure[Study of OM in \texttt{PP-3v1}.]{\includegraphics[width=.32\textwidth]{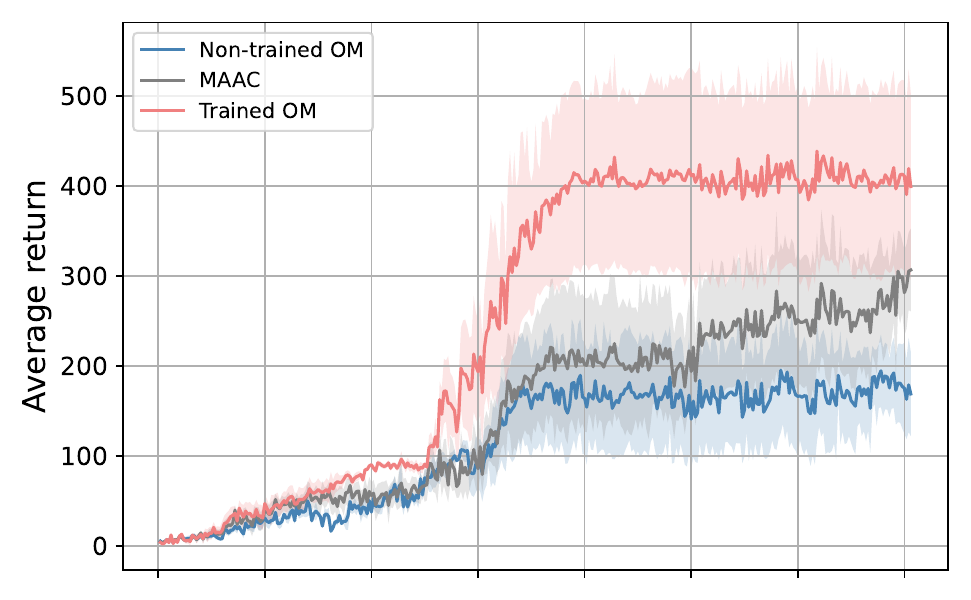}\label{fig:ablation studie of of trained model in PP-4v2}}
    \subfigure[Study of OM  in \texttt{Pomm-FFA}.]{\includegraphics[width=.32\textwidth]{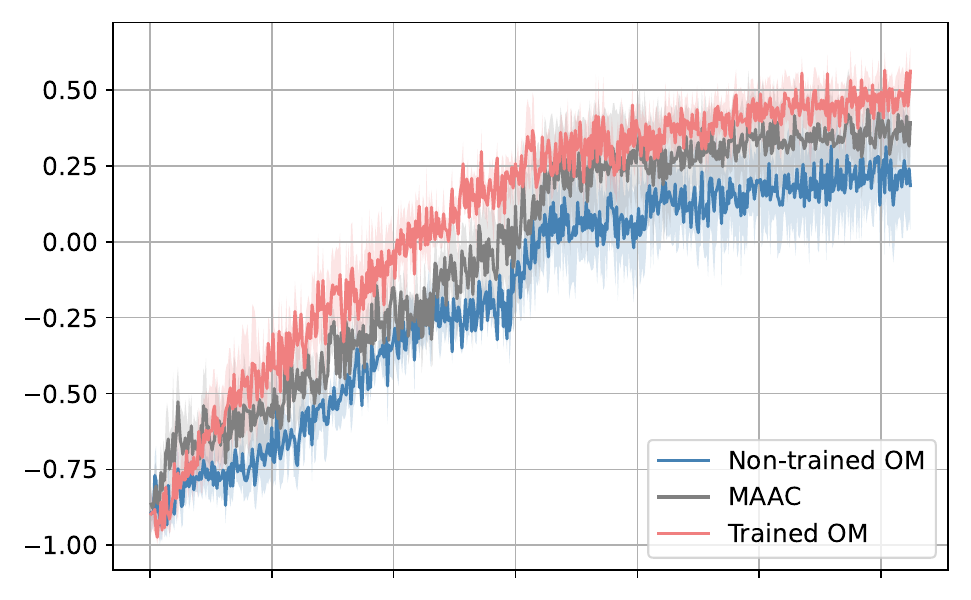}\label{fig:ablation studie of of trained model in Pomm-FFA}} 
    \subfigure[Study of OM in \texttt{Pomm-Team}.]{\includegraphics[width=.32\textwidth]{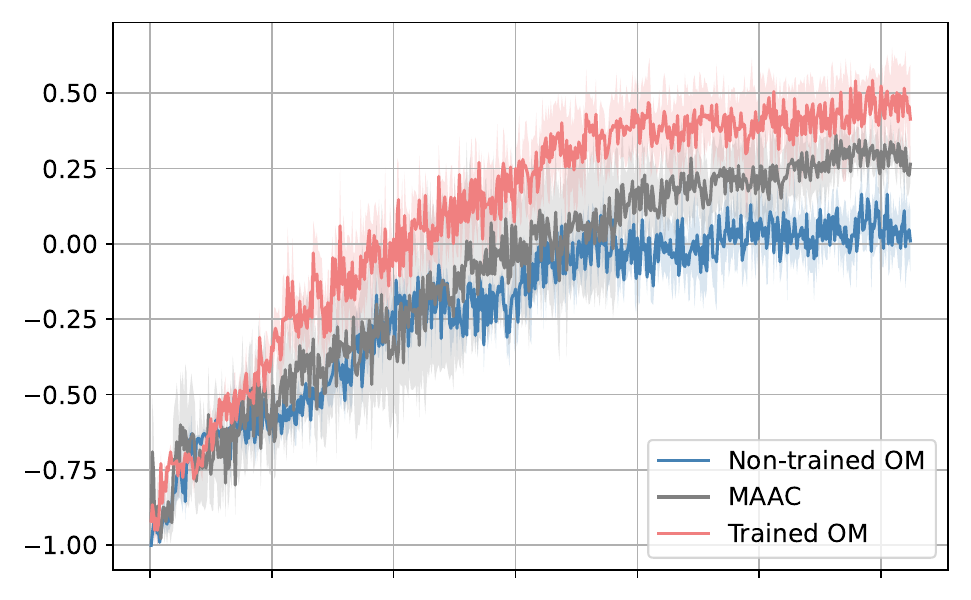}\label{fig:ablation studie of of trained model in Pomm-Team}}\\
    \subfigure[OM accuracy for \texttt{PP-3v1}. ]{\includegraphics[width=.32\textwidth]{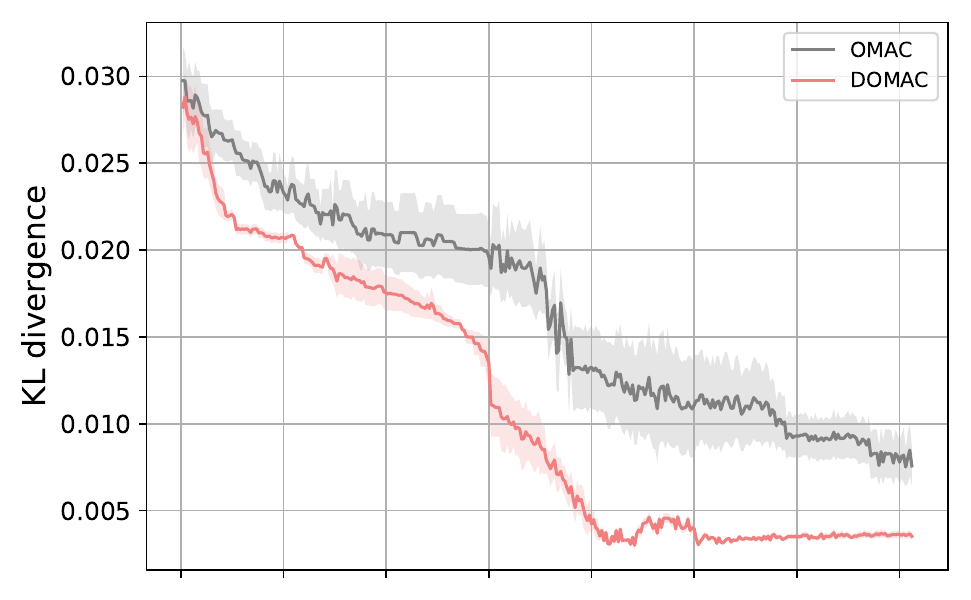}\label{fig:OM accuracy in PP2v1}} \vspace{-1mm}
    \subfigure[OM accuracy for  \texttt{Pomm-FFA}.]{\includegraphics[width=.32\textwidth]{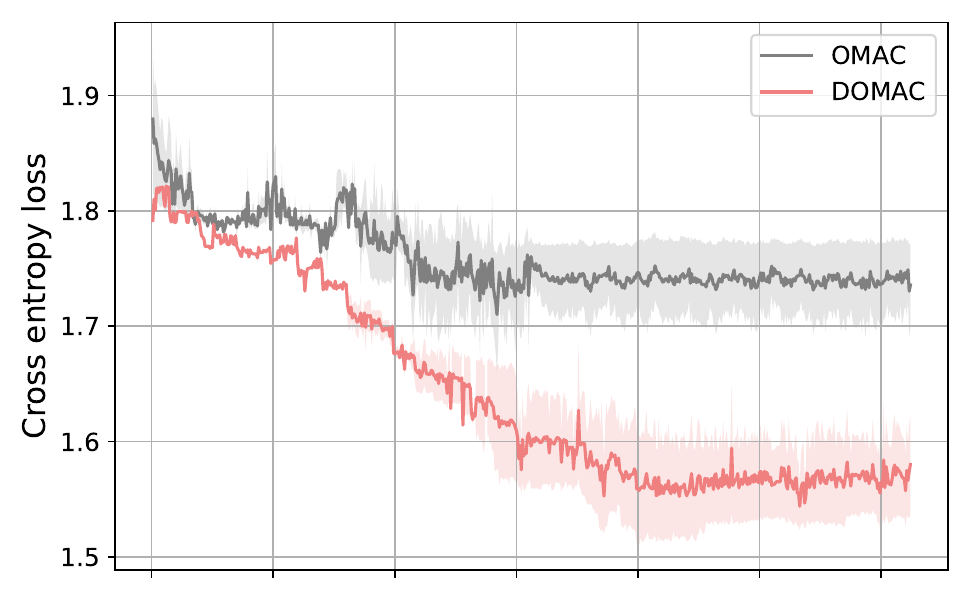}\label{fig:OM accuracy in Pomm-FFA}} \vspace{-1mm}
    \subfigure[OM accuracy for \texttt{Pomm-Team}.]{\includegraphics[width=.32\textwidth]{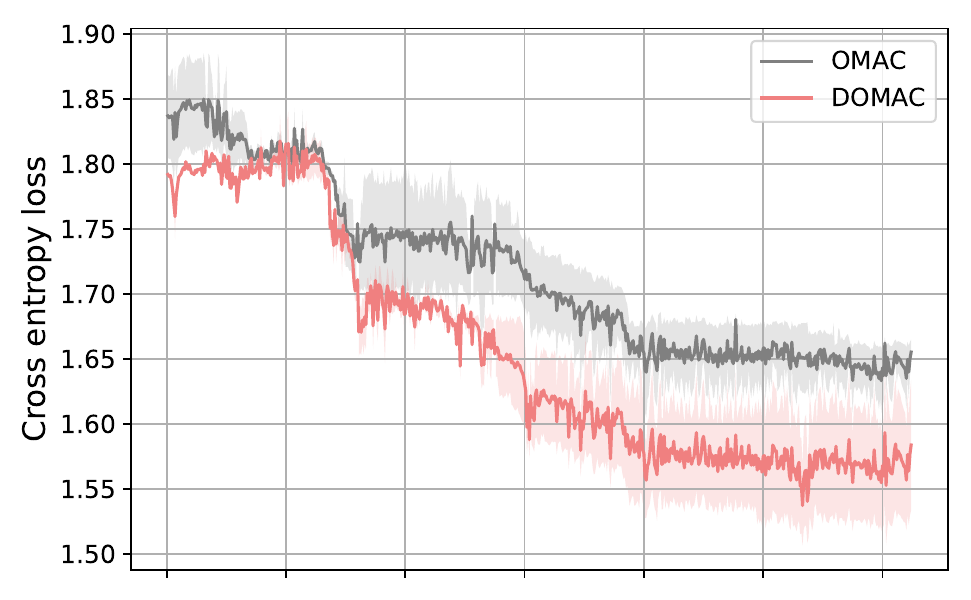}\label{fig:OM accuracy in Pomm-Team}} \vspace{-1mm} \\
     \subfigure[Entropy of OM for \texttt{PP-3v1}.]{\includegraphics[width=.32\textwidth]{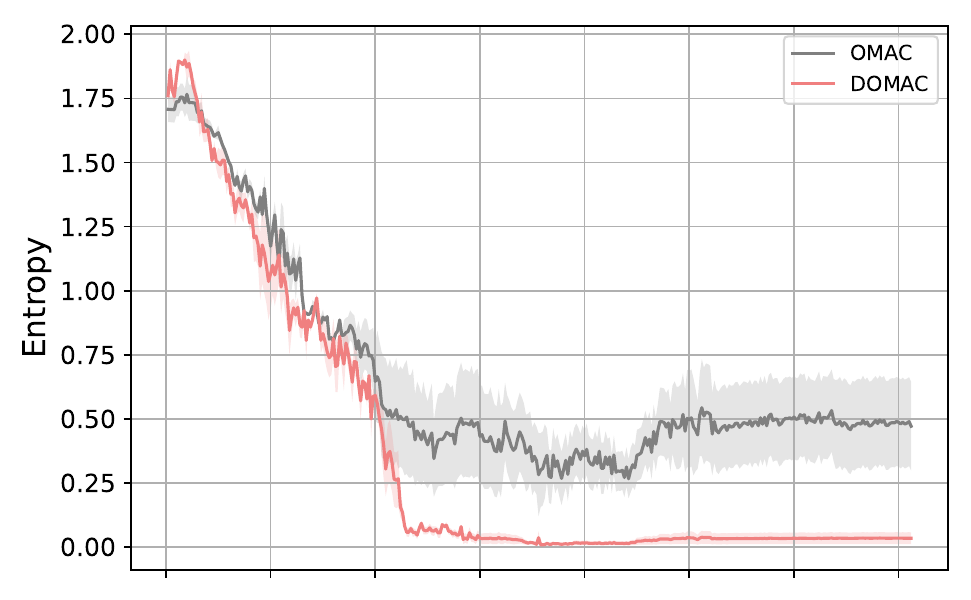}\label{fig:entropy of OM in PP3v1}}
     \subfigure[Entropy of OM for \texttt{Pomm-FFA}.]{\includegraphics[width=.32\textwidth]{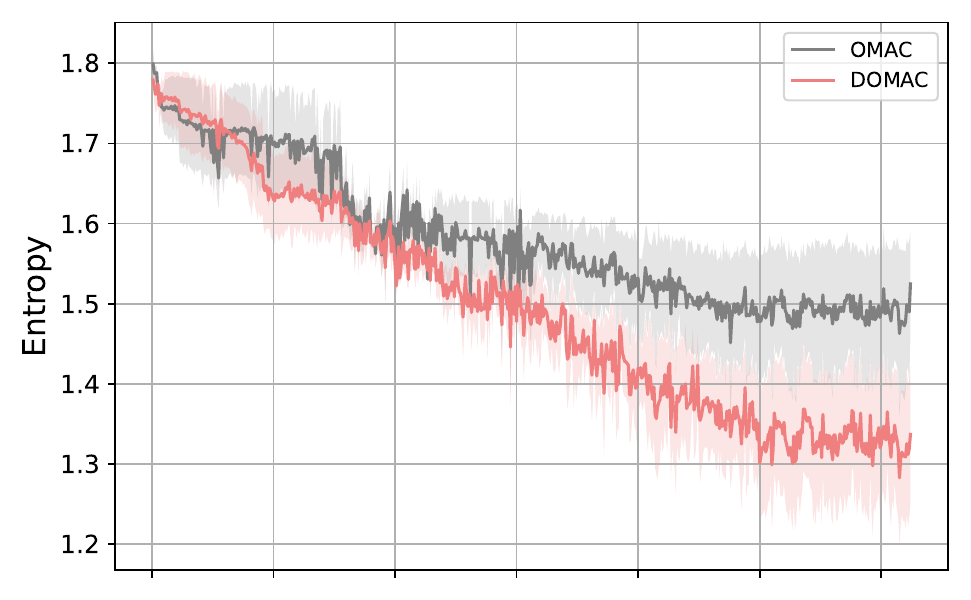} \label{fig:entropy of OM in Pomm-FFA}}
     \subfigure[Entropy of OM for \texttt{Pomm-Team}.]{\includegraphics[width=.32\textwidth]{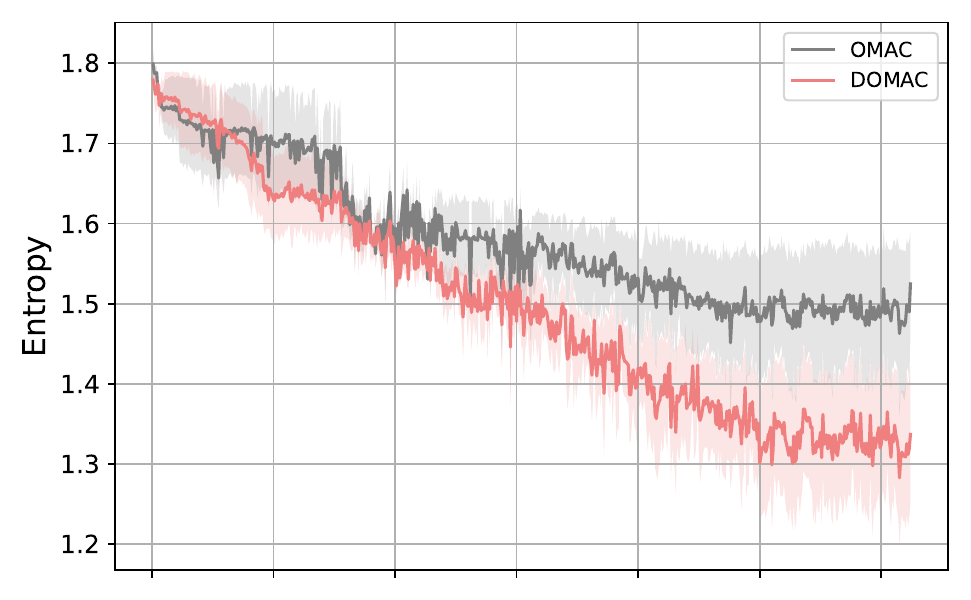}\label{fig:entropy of OM in Pomm-Team}}\\
    \subfigure[Study for K in \texttt{PP-3v1}.]{\includegraphics[width=.32\textwidth]{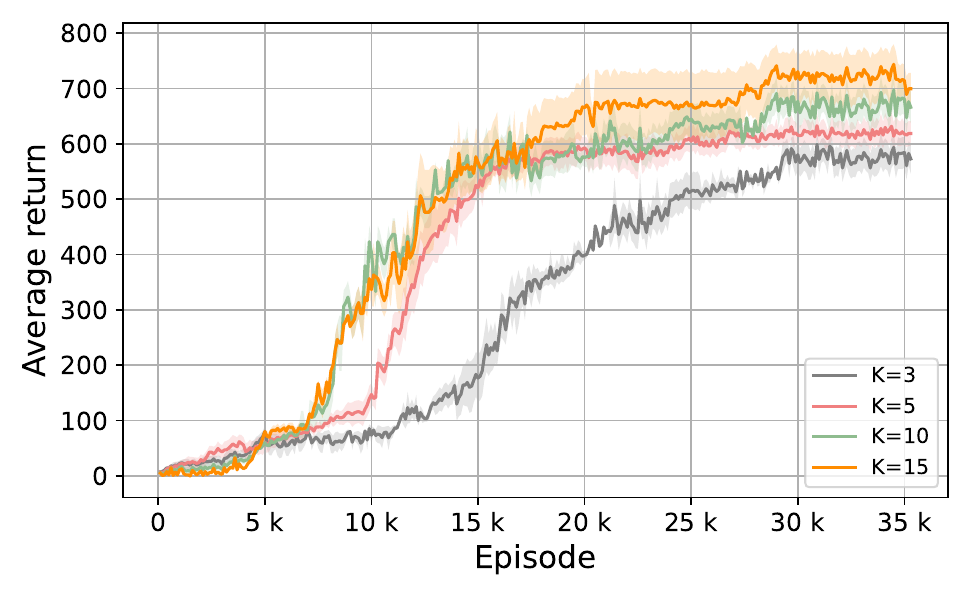}\label{fig: ablation studie of K in PP3v1}}
    \subfigure[Study for K in \texttt{Pomm-FFA}.]{\includegraphics[width=.32\textwidth]{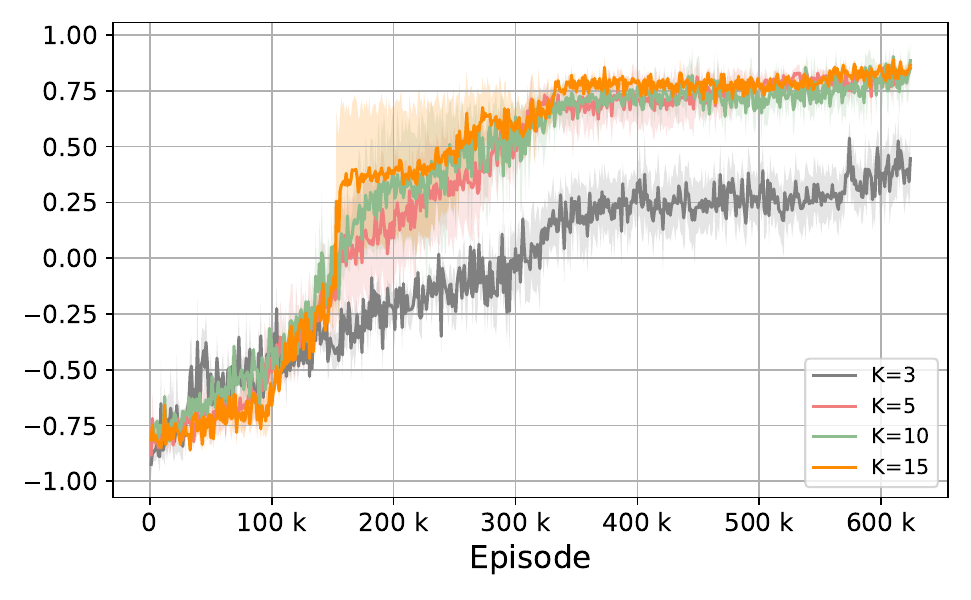}\label{fig: ablation studie of K in Pomm-FFA}} 
    \subfigure[Study for K in \texttt{Pomm-Team}.]{\includegraphics[width=.32\textwidth]{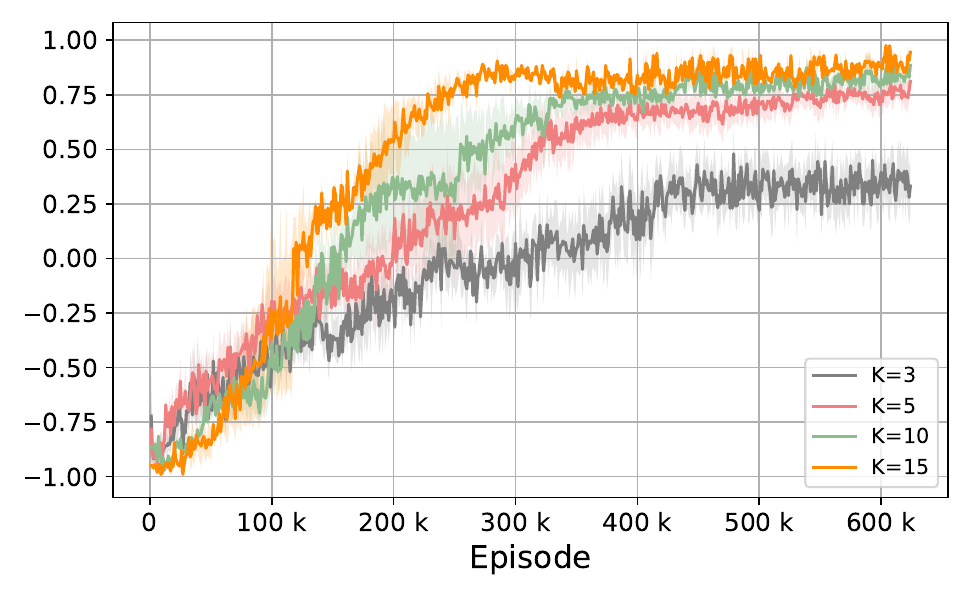}\label{fig: ablation studie of K in Pomm-Team}}
\caption{Ablation studies for opponent models (OM) and quantile number $K$ in $\texttt{PP-3v1}$, $\texttt{Pomm-FFA}$, $\texttt{Pomm-Team}$.}
\label{figa1}
\end{figure*}

\renewcommand{\thefigure}{A2}
\begin{figure*}[!t]
\footnotesize
    \centering
    \subfigure[\texttt{3v1}]{\includegraphics[width=.32\textwidth]{{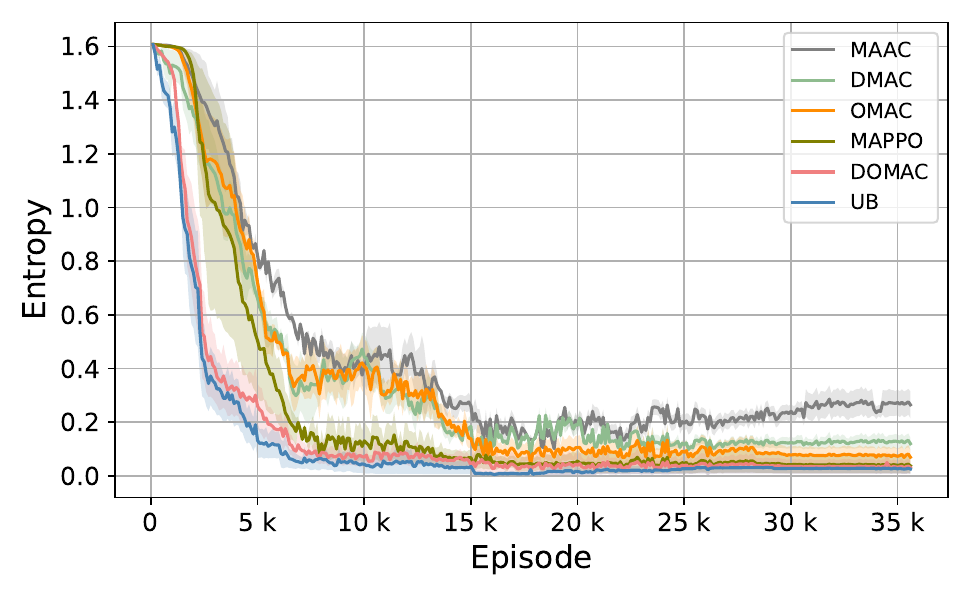}}\label{3v1}}
    \subfigure[\texttt{FFA}]{\includegraphics[width=.32\textwidth]{{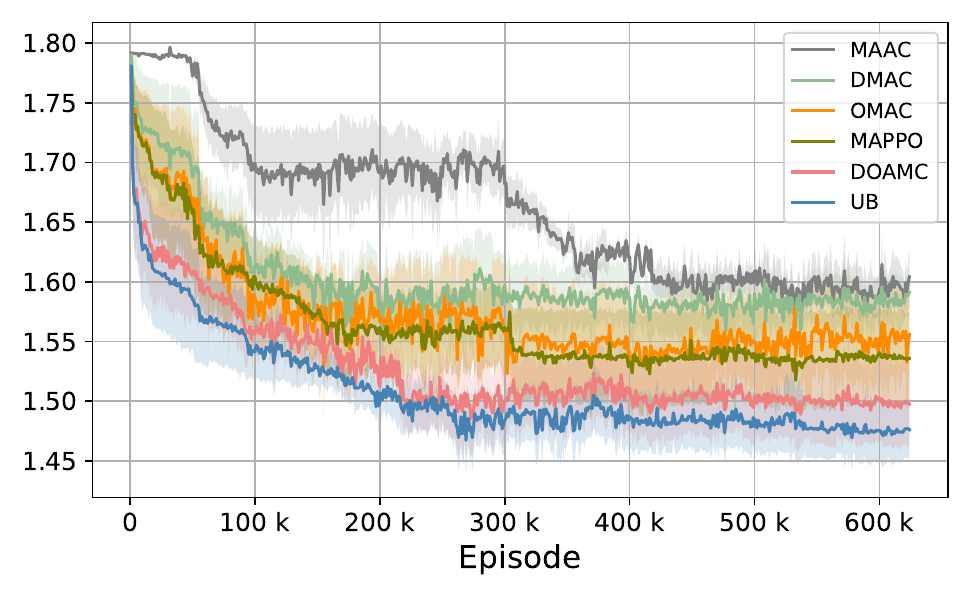}}\label{FFA}}
    \subfigure[\texttt{2s3z}]{{\includegraphics[width=.32\textwidth]{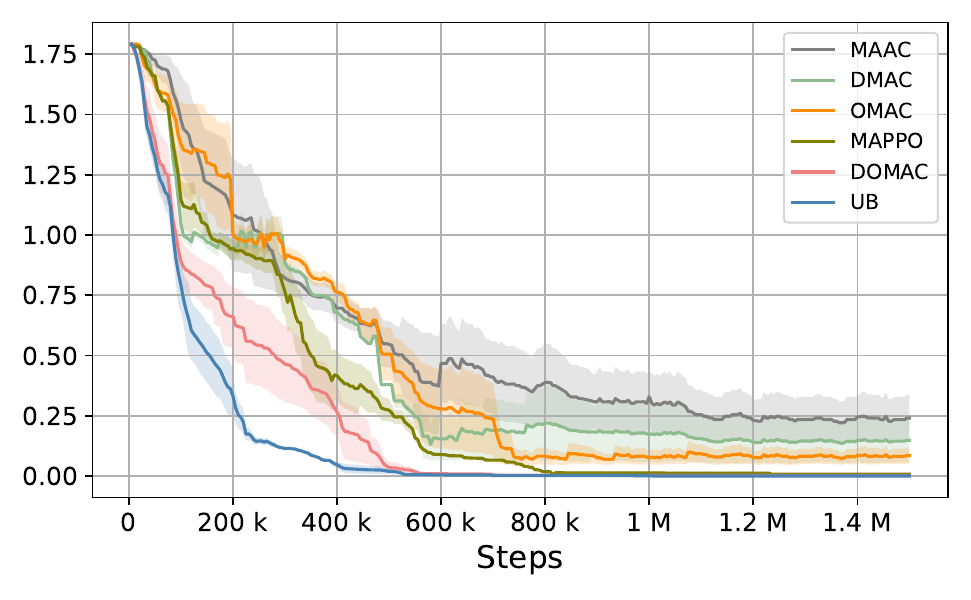}}\label{2s3z}}
    \caption{{{Average entropy of policies for game \texttt{FFA} and \texttt{2s3z}.}}}
    \label{figa2}
\end{figure*}

\end{document}